\documentclass[11pt,a4paper]{article}
\usepackage[blocks]{authblk}
\usepackage[hyperref]{acl2019}
\usepackage{times}
\usepackage{latexsym}

\usepackage{url}

\usepackage{multirow}
\usepackage{amsmath}
\usepackage{amsfonts}
\usepackage{graphicx}
\usepackage{footnote}
\usepackage{subcaption}

\aclfinalcopy %

\title{Augmenting Neural Response Generation with Context-Aware Topical Attention}

\author[ ]{Nouha Dziri}
\author[ ]{Ehsan Kamalloo}
\author[ ]{Kory W. Mathewson}
\author[ ]{Osmar Zaiane}
\affil[ ]{Department of Computing Science}
\affil[ ]{University of Alberta}
\affil[ ]{
{\tt \{dziri,kamalloo,korym,zaiane\}@cs.ualberta.ca}}

\date{}

\begin{document}
\maketitle
\begin{abstract}
 Sequence-to-Sequence (Seq2Seq) models have witnessed a notable success in generating natural conversational exchanges. Notwithstanding the syntactically well-formed responses generated by these neural network models, they are prone to be acontextual, short and generic. In this work, we introduce a Topical Hierarchical Recurrent Encoder Decoder (THRED), a novel, fully data-driven, multi-turn response generation system intended to produce contextual and topic-aware responses.
Our model is built upon the basic Seq2Seq model by augmenting it with a hierarchical joint attention mechanism that incorporates topical concepts and previous interactions into the response generation.
To train our model, we provide a clean and high-quality conversational dataset mined from Reddit comments. We evaluate THRED on two novel automated metrics, dubbed Semantic Similarity and Response Echo Index, as well as with human evaluation. 
Our experiments demonstrate that the proposed model is able to generate more diverse and contextually relevant responses compared to the strong baselines.
\end{abstract}

\section{Introduction} \label{intro}
With the recent success of deep neural networks in natural language processing tasks such as machine translation \cite{sutskever2014sequence} and language modeling \cite{mikolov2010recurrent}, there has been growing research interest in building data-driven dialogue systems.
Fortunately, innovation in deep learning architectures and the availability of large public datasets have produced fertile ground for the data-driven approaches to become feasible and quite promising. In particular, the Sequence-to-Sequence (Seq2Seq) neural network model \cite{sutskever2014sequence} has witnessed substantial breakthroughs in improving the performance of conversational agents.
Such a model succeeds in learning the backbone of the conversation but lacks any aptitude for producing context-sensitive and diverse conversations.
Instead, generated responses are dull, short and carry %
little information \cite{li2015diversity}.
Instinctively, humans tend to adapt conversations to their interlocutor not only by looking at the last utterance but also by considering information and concepts covered in the conversation history \cite{danescu2011chameleons}. Such adaptation 
increase the smoothness and engagement of the generated responses. 
We speculate that incorporating conversation history and topic information with our novel model and method will improve generated conversational responses.
In this work, we introduce a novel, fully data-driven,  multi-turn response generation system intended to produce context-aware and diverse responses.
Our model builds upon the basic Seq2Seq model by combining conversational data and external knowledge information trained through a  hierarchical joint attention neural model.
We find that our method leads to both diverse and contextual responses compared to the literature strong baselines. We also introduce  two novel quantitative metrics for dialogue model development, dubbed  Semantic Similarity and Response Echo Index. While the former measures 
  the  capability  of  the  model  to  be  consistent  with  the  context  and  to maintain the  topic  of  the  conversation,  the  latter  assesses  how  much  our approach  is  able to  generate  unique and plausible  responses which are measurably distant from the input
dataset.
Used together, they provide a means to reduce burden of human evaluation and
 allow rapid testing of dialogue models. We show that such metrics correlate well with human judgment, making a step towards a good automatic evaluation procedure.

The key contributions of this work are:
\begin{itemize}
  
    \item We devise a fully data-driven neural conversational model that leverages conversation history and topic information in the response generation process through a hierarchical joint attention mechanism; making the dialogue more diverse and engaging.
    \item We introduce two novel automated metrics: Semantic Similarity and Response Echo Index and we show that they correlate well with human judgment.

    \item We collect, parse and clean a conversational dataset from Reddit comments\footnote{The source code and the dataset are available at \url{https://github.com/nouhadziri/THRED}}.

\end{itemize}

\section{Related Work}

Neural generative models have been improved through several techniques. \cite{serban2016building} built upon the Seq2Seq work  by introducing a Hierarchical Recurrent Encoder-Decoder neural network (HRED) that accounts for the conversation history.
\cite{li2016deep} used deep reinforcement learning to generate highly-rewarded responses by considering three dialogue properties: ease of answering, informativeness and coherence. \cite{zhang2018personalizing} addressed the challenge of personalizing the chatbot by modeling human-like behaviour. They presented a persona-based model that aims to handle the speaker consistency by integrating a speaker profile vector representation into the the Seq2Seq model.
\cite{xing2017topic} used a similar idea but added an extra probability value in the decoder to bias the overall distribution towards leveraging topic words in the generated responses. Their architecture does not focus on capturing conversation history. All of these improvements are motivated by the scarcity of diversity and informativeness of the responses. Our work follows on from these works with the additional aim of generating context-aware responses by using a hierarchical joint attention model.
An important line of research that we also address in this work is automatically evaluating the quality of dialogue responses. In dialogue systems, automated metrics tend to be borrowed from other NLP tasks such as BLEU \cite{papineni2002bleu} from machine translation and
ROUGE \cite{lin2004rouge} from text summarization.
Yet, such metrics fail, mainly because they are focusing on the word-level overlap between the machine-generated answer and the human-generated answer, which can be inconsistent with what humans deem a plausible and interesting response. \cite{liu2016not} have showed that these metrics correlate very weakly with human evaluation. Indeed, word-overlapping metrics achieve best results when the space of responses is small and lexically overlapping which is not the case for dialogue systems responses. 
Significant works have looked into this challenge. Examples include ADEM \cite{lowe2017towards}, an evaluation model that learns to score responses from an annotated dataset of human responses scores.
\cite{venkatesh2018evaluating} proposed a number of metrics based on user experience, coherence, and topical diversity and have showed that these metrics can be used as a proxy for human evaluation. However, engagement and coherence metrics are estimated via recruiting evaluators. In this work, we propose directly calculable approximations of human evaluation grounded in conversational theories of accommodation and affordance \cite{danescu2011chameleons}.

\section{Topical Hierarchical Recurrent Encoder Decoder}
\begin{figure*}[t]
\centering
\includegraphics[scale=0.17]{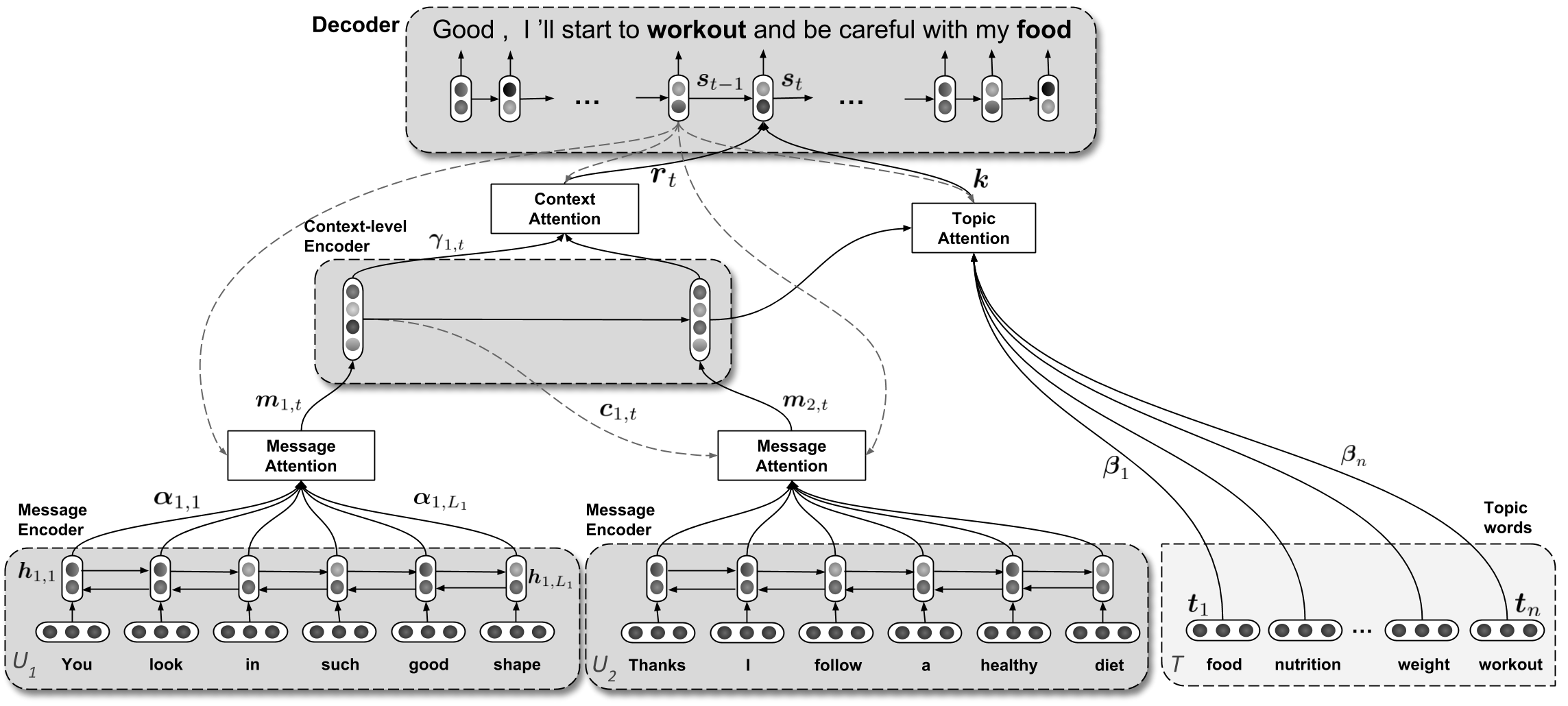}
\caption{THRED model architecture in which we jointly model two specifications that presumably make the task of response generation successful: context-awareness (modeled by \textbf{Context Attention}) and diversity (modeled by \textbf{Topic Attention}).}
\label{Figure 1: seriea}
\end{figure*}

Topical Hierarchical Recurrent Encoder Decoder (THRED) can be viewed as a hybrid model that conditions the response generation on conversation history captured from previous utterances and on topic words acquired from a Latent Dirichlet Allocation (LDA) model \cite{blei2003latent}. The proposed approach extends the standard Seq2Seq model by leveraging topic words in the process of response generation and accounting for conversation history. Figure \ref{Figure 1: seriea} illustrates our model.
We detail below the components of our model.

\subsection{Message Encoder}
Let a sequence of $N$ utterances within a dialogue $D= \{U_1,..., U_N\}$. Every utterance $U_i=\{w_{i,1}, ..., w_{i,L_i}\}$ contains a random variable $L_i$ of sequence of words where $w_{i,k}$ represents the word embedding vector at position $k$ in the utterance $U_i$.
The message encoder sequentially accepts the embedding of each word in the input message $U_i$ and updates its hidden state at every time step $t$ by a bidirectional GRU-RNN \cite{cho2014learning} according to:
\setlength{\abovedisplayskip}{1pt}
 \begin{equation}\label{gru encoder}
    h_{i,t} = GRU(h_{i,t-1}, w_{i,t}),  \forall t \in \{1, \ldots ,L_i\} 
\end{equation}
\setlength{\belowdisplayskip}{1pt}
where $h_{i,t-1}$ represents the previous hidden state.

\subsection{Message Attention} 
Different parts of the conversation history have distinct levels of importance that may influence the response generation process. The message attention in THRED operates by putting more focus on the salient input words with regard to the output. 
It computes, at step $t$, a weight value $\alpha_{i,j,t}$ for every encoder hidden state $h_{i,j}$ and linearly combines them to form a vector \textbf{$m_{i,t}$} according to Bahdanau attention mechanism \cite{bahdanau2014neural}. Formally, $m_{i,t}$ is calculated as:
\setlength{\abovedisplayskip}{1pt}
 \begin{equation}\label{context attention}
    m_{i,t} = \sum_{j=1}^{L_i}\alpha_{i,j,t} \; h_{i,j}, \: \forall i \in \{1, \ldots ,N\}
\end{equation}
\setlength{\belowdisplayskip}{1pt}
where $\alpha_{i,j,t}$ is computed as:
\setlength{\abovedisplayskip}{1pt}
 \begin{equation*}\label{attention}
    \begin{split}
    \alpha_{i,j,t} &= \frac{\exp(e_{i,j,t})}{\sum_{k=1}^{L_i} \exp(e_{i,k,t})} \: ; \\
    e_{i,j,t} &= \eta(s_{t-1}, h_{i,j}, c_{i-1,t})
    \end{split}
\end{equation*}
\setlength{\belowdisplayskip}{1pt}

where $s_{t-1}$ represents the hidden state of the decoder (further details are provided later), $c_{i,t}$ delineates the hidden state of the context-level encoder (computed in Equation \eqref{eq:contextRNN}) , $\eta$ is a multi-layer perceptron having tanh as activation function. Unlike the Bahdanau attention mechanism, the attentional vector \textbf{$m_{i,t}$} is based on both the hidden states of the decoder and the hidden states of the context-level encoder. We are motivated by the fact that $c_{i,t}$ may carry important information that could be missing in $s_{t-1}$.
In summary, the attentional vector \textbf{$m_{i,t}$} is an order-sensitive information of all the words in the sentence, attending to more important words in the input messages.

\subsection{Context-Level Encoder}
The context-level encoder takes as input each utterance representation ($m_{1,t}$, \ldots , $m_{N,t}$) and calculates the sequence of recurrent hidden states as shown in Equation \eqref{eq:contextRNN}:

\begin{equation} \label{eq:contextRNN}
    c_{i,t} = GRU(c_{i-1,t}, m_{i,t}),  \forall i \in \{1, \ldots ,N\}
\end{equation}

where $c_{i-1,t}$ delineates the previous hidden state of the context-level encoder and N represents the number of utterances in the conversation history.
The resulted $c_{i,t}$ vector summarizes all past information that have been processed up to position $i$.

\subsection{Context-Topic Joint Attention}

\textbf{Context Attention:}
On top of the context-level encoder, a context attention is added to attend to important utterances in the conversation history. Precisely, the context attention assigns weights $(\gamma_{1,t}, ..., \gamma_{N,t})$ to $(c_{1,t}, ..., c_{N,t})$ and forms a vector $r_t$ as
\vspace{-1em}
 \begin{equation}\label{context attention hier}
    r_t = \sum_{j=1}^{N}\gamma_{j,t} c_{j,t}
\end{equation}
\vspace{-1em}

where:
 \begin{equation}\label{attention}
    \begin{split}
        \gamma_{j,t} &= \frac{\exp(e'_{j,t})}{\sum_{i=1}^{N} \exp(e'_{i,t})} \: ; \\
        e'_{i,t} &= \eta(s_{t-1}, c_{i,t})
    \end{split}
\end{equation}

\textbf{Topic Attention:} 
In order to infuse the response with information relevant to the input messages, we enhance the model with topic information.
We assign a topic $T$ to the conversation context using a pre-trained LDA model \cite{hoffman2010online}. LDA is a probabilistic topic model that appoints multiple topics for the dialogue history.
The LDA parameters were estimated using the collapsed Gibbs sampling algorithm \cite{zhao2011comparing}. We provide further details on how we train this model in the supplementary material. In our case, the conversation history is a short document, so we believe that the most probable topic will be sufficient to model the dialogue. After acquiring topic words for the entire history, we pick the $n$ highest probable words under $T$ (we choose $n=100$ in our experiments). 
The topic words $\{t_1, \cdots ,t_n\}$ are then linearly combined to form a fixed-length vector \textbf{$k$}. The weight values are calculated as the following:
\setlength{\abovedisplayskip}{1pt} 
 \begin{equation}\label{eq:topic}
 \beta_{i,t} = \frac{\exp(\eta(s_{t-1}, t_i , c_{N,t}))}{\sum_{j=1}^{n} \exp(\eta(s_{t-1}, t_j ,c_{N,t}))}
\end{equation}
\vspace{-.6em}

where $i \in \{1, \cdots ,n\}$, $c_{N,t}$ is the last hidden state of the context-level encoder, and $s_{t-1}$ is the ${t-1}^{th}$ hidden state in the decoder. 
The topic attention uses additionally the last hidden state of the context-level encoder $c_{N,t}$ in order to diminish the repercussion of impertinent topic words and feature the relevant ones to the message. Unlike \cite{xing2017topic}, our model employs the final context-level encoder hidden state $c_{N,t}$ in order to account for conversation history in the generated response.
In summary, the topic words are summarized as a topic vector $k$ representing prior knowledge for response generation. The key idea of this approach is to affect the generation process by
avoiding the need to learn the same conversational
pattern for each utterance but instead enriching the responses with topics and words related to the subject of the message even if the words were never used before.

\vspace{-0.4em}
\subsection{Decoder}
The decoder is responsible for predicting the response utterance $U_{m+1}$ given the previous utterances and the topic words. Following \cite{xing2017topic}, we biased the generation probability towards generating the topic words in the response. In particular, we added an extra probability to the standard generation probability, enforcing the model to account for the topical tokens. Consequently, the generation probability is defined as the following: 
\begin{equation}
p(w_i) = p_V(w_i) + p_K(w_i)
\end{equation}
where $K$ and $V$ represent respectively topic vocabulary and response vocabulary; $p_V$ and $p_K$ are defined as follows:
\begin{equation*}
    \begin{split}
  p_V(w_i) &= \frac{1}{M} \exp (\sigma_V(s_i, w_{i-1})) \\
  p_K(w_i) &= \frac{1}{M} \exp (\sigma_K(s_i, w_{i-1}, r_i))
    \end{split}
\end{equation*}

where $s_i = GRU(w_{i-1}, s_{i-1}, r_i, k )$, $\sigma$ is a tanh and  M is calculated as follows:

\[
\begin{split}
M &= \sum_{v \in V} \exp{(\sigma_V(s_i, w_{i-1}))} \\
&+ \sum_{v' \in K} \exp{(\sigma_K(s_i, w_{i-1}, r_i))}
\end{split}
\]
\vspace{-1em}

\section{Datasets} \label{sec:datasets}

One of the main weaknesses of dialogue systems is caused by the paucity of high-quality conversational dataset. The well-known OpenSubtitles dataset \cite{tiedemann2012parallel} lacks speaker annotations, thus making it more difficult to train conversation systems which demand high quality speaker and conversation level tags. Therefore, the assumption of treating consecutive utterances as turn exchanges uttered by two persons \cite{vinyals2015neural} could not be viable.
To enable the study of high-quality and large-scale dataset for dialogue modeling, we have collected a corpus of 35M conversations drawn from the Reddit data\footnote{\url{https://files.pushshift.io/reddit/}}, where each dialogue is composed of three turn exchanges. 
The Reddit dataset is composed of posts and comments, 
where each comment is annotated with rich meta data (i.e., author, number of replies, user's comment karma, etc.)\footnote{\url{https://github.com/reddit-archive/reddit/wiki/JSON}}. To harvest the dataset, we curated 95 English subreddits out of roughly 1.1M public subreddits\footnote{As of February 2019}.
Our choice was based on the top-ranked subreddits that discuss topics such as news, education, business, politics and sports. We processed Reddit for a 12 month-period ranging from December 2016 until December 2017. For each post, we retrieved all comments and we recursively followed the chain of replies of each comment to recover the entire conversation. Reddit dataset is often semantically well-structured and is not filled with spelling errors thanks to moderator's efforts. Therefore, we do not perform any spelling correction procedure.
Due to resource limitations, we randomly sampled 6M dialogues as training data, 700K dialogues as development data, and 40K dialogues as test data. For OpenSubtitles, we trained the models on the same size of data as for Reddit.

\begin{table*}[t]
\linespread{0.8}\selectfont\centering
\centering
\begin{tabular}{l|p{9.5cm}} 
 	\hline
    {\small \textsc{Context}} & {\small \textsc{Generated Responses}} \\
    \hline
        \multirow{1}{5.5cm}{{\footnotesize \textbf{(Reddit)} sanctions are an act of war \textcolor{blue}{\textbf{$\rightarrow$}}  why do you think that ?}}
       
       & {\footnotesize \textbf{THRED:} because it's really a \textcolor{red}{theory} that supports \textcolor{red}{terrorism} . and this has an effect on the idea of a \textcolor{red}{regime} that isn't the same as a \textcolor{red}{government} (\underline{{\em Excellent, Good, Excellent, Good, Excellent}})} \\
       & {\footnotesize \textbf{HRED:} because the war is n't a war . it 's a war . (\underline{{\em Good, Poor, Poor, Poor, Poor}})} \\
       & {\footnotesize \textbf{Seq2Seq:} 
because it 's an unpopular opinion , and that 's why it 's a bad thing to say .
 (\underline{{\em Good, Poor, Excellent, Good, Good}})} \\
       & {\footnotesize \textbf{TA-Seq2Seq:} because it's a war . (\underline{{\em Good, Poor, Excellent, Poor, Good}})} \\
    \hline
        
\end{tabular}
\linespread{1.0}
\caption{One cherry-picked dialogue out of 150 conversations along with the generated responses from all models. Human judgments are provided in the brackets. 
The blue arrow specifies a dialogue exchange and the highlighted words in red represent the topic words acquired from the pre-trained LDA model. }
\label{tab:SampleResponsesReddit}
\end{table*}

\section{Experiments}

In this section, we focus on the task of evaluating the next utterance given the conversation history. We compare THRED against three open-source baselines, namely Standard Seq2Seq with attention mechanism \cite{bahdanau2014neural}, HRED \cite{serban2016building}, and Topic-Aware (TA) Seq2Seq \cite{xing2017topic}. As done in \cite{li2016deep}, for Standard Seq2Seq and TA-Seq2Seq, we concatenate the dialogue history to account for context in a multi-turn conversation. All experiments are conducted on two datasets (i.e., Reddit and OpenSubtitles). We report results on OpenSubtitles in the supplementary material.

\subsection{Quantitative Evaluation}

In the following subsections,
 we introduce two metrics that can impartially evaluate THRED and compare against the different baselines.
These metrics were tested on 5000 dialogues randomly sampled from the test dataset.
It is worth mentioning that we present word perplexity (PPL) on the test data in Table~\ref{tab:diversity} (along with diversity metric). However, we do not believe that it represents a good measure for assessing the quality of responses \cite{serban2017hierarchical}. This is because perplexity captures how likely the  responses are under a generation probability distribution, and does not measure the degree of diversity and engagingness in the responses. 

\subsection{Semantic Similarity}
A good dialogue system should be capable of sustaining a coherent conversation with a human by staying on topic and by following a train of thoughts \cite{venkatesh2018evaluating}. 
Semantic Similarity (SS) metric estimates the correspondence between the utterances in the context and the generated response. The intuition behind this metric is that plausible responses should be consistent
with the context and should maintain the topic of the conversation.
Our response generator THRED along with the baselines generate an utterance based on the two previous
utterances in the dialogue (i.e., Utt1 and Utt2). We compute the cosine distance between the embedding vectors of the test utterances (Utt.1 and
Utt.2) and the generated responses from the different models (i.e., THRED, TA-Seq2Seq, HRED and Seq2Seq).  Therefore, a low score denotes a high coherence.
More precisely,  for each triple in the test dataset, we test two scenarios: (1) we compute the SS of each generated response with respect to the most recent utterance in the conversation (Utt.2) and (2) we compute the SS of each generated response with respect to the second most recent utterance (Utt.1). 
To render the semantic representation of an utterance, we leverage Universal Sentence Encoder \cite{cer2018universal} wherein a sentence is projected to a fixed dimensional embedding vector.

However, dull and generic responses such as ``{\em i'm not sure}" tend to be semantically close to many utterances, hindering the effectiveness of the metric. To cope with this negative effect, we manually compiled a set of 60 dull responses and computed the SS score by multiplying the cosine distance with the following penalty factor (akin to length penalty in \cite{wu2016google}):

\vspace{-1em}
\begin{equation*}
    P = 1 + \log \frac{2 + L'}{2 + L''}
\end{equation*}
 \vspace{-1em}
 
where $L'$ indicates the length of the response after dropping stop words and punctuation and $L''$ stands for the length of non-dull part of the response after dropping stop words. The intuition here is that the longer utterances, with nearly the same similarity, communicate the intention unequivocally since it takes more words to convey the same meaning.\\
The penalized Semantic Similarity score  is therefore defined as:
\begin{equation*}
    SS(utt_{i,j} \: , \: resp_i) = P \times (1- cos(\vec{utt_{i,j}}, \vec{resp_i}))
\end{equation*}
where $i$ represents the index for the dialogue in the test dataset and $j$ denotes the index of the utterance in the conversation history.
\begin{table}[t]
\centering
\begin{tabular}{ l c c c c }
 \hline
  \small \textbf{Stat.} & \small \textbf{THRED} & \small \textbf{Seq2Seq} & \small \textbf{HRED} & \small \textbf{TA-Seq2Seq} \\
 \hline
 \multicolumn{5}{l}{{\small SS with respect to Utt.1}} \\
$\mu$ & \small \textbf{0.680} & \small 0.694 & \small 0.755 & \small 0.692\\
$\sigma$ & {\small 0.200} & {\small 0.236} & {\small 0.283} & {\small 0.252} \\ 
\multicolumn{5}{l}{{\small SS with respect to Utt.2}} \\
$\mu$ & {\small \textbf{0.649**}} & \small 0.672 &\small 0.720 & \small 0.702 \\
$\sigma$ & {\small 0.212} & \small 0.236 & \small 0.292 & \small 0.253 \\
 \hline
\end{tabular}
\caption{{\small Mean $\mu$ and standard deviation $\sigma$ of SS scores for the responses generated from different models with respect to the most recent utterance (Utt.2) and the 2nd most recent utterance (Utt.1) from conversation history on the Reddit test dataset (\textbf{**} indicates statistical significance over the second best method with $p$-value $<$ 0.001).}}  
\label{tab:SC}
\end{table}
The results conducted on Reddit dataset are shown in Table \ref{tab:SC}.
We can observe that THRED is able to generate responses which follow the topic and semantics of the input utterances. In particular, the responses generated by THRED tend to be closer to the context of the conversation (Utt.1 and Utt.2) compared to the responses generated from the baslines. To ensure the statistical significance of THRED, we conducted Student's $t$-test over the average values of SS metric. THRED outperforms all baselines ($p<0.001$) especially when the comparison is made against the most recent utterance (Utt.2).
On the other hand, THRED is level with compared models in semantic distance with respect to the second most recent utterance (Utt.1). This makes sense because in a multi-turn dialogue, speakers are more likely to address the last utterance spoken by the interlocutor, which is why THRED tends to favour the most recent utterance over an older one. Additionally, the roughly similar distances for both utterances in Standard Seq2Seq and TA-Seq2Seq exhibit that by concatenating context as single input, these models cannot distinguish between early turns and late turns.
 Similarly, the results achieved on OpenSubtitles dataset (See Figure \ref{fig:semantic_similarity} in the supplementary material) illustrate that THRED succeeds in staying on topic and in  accounting for contextual information.

\subsubsection{Reliability Assurance}
In order to ensure that the SS measurement is stable and void of random error, we investigate whether the SS metric  is able to yield the same previous results regardless of a specific test dataset. Following \cite{papineni2002bleu}, the test dataset is randomly partitioned to 5 disjoint subsets (i.e., each one consists of 1000 test dialogues). Then, we compute standard deviation of SS over each dataset. The results, showcased in Table~\ref{tab:Reliability}, indicate low standard deviation on the subdatasets, denoting that the SS metric is a  consistent and reliable measure to compare different dialogue models.

\begin{table}[t]
\centering
\begin{tabular}{ l | c c c c }
\hline
  {\small \textbf{Metric}} & {\footnotesize \textbf{THRED}} & {\footnotesize \textbf{Seq2Seq}} & {\footnotesize \textbf{HRED}} & {\footnotesize \textbf{TA-Seq2Seq}} \\
 \hline
 {\small SS$_{\text{w.r.t. Utt.1}}$} & {\small 0.008}	& {\small 0.009}	& {\small 0.001}	& {\small 0.006} \\
 {\small SS$_{\text{w.r.t. Utt.2}}$} & {\small 0.010}	& {\small 0.008}	& {\small 0.007}	& {\small 0.005} \\
 \hline
\end{tabular}
\caption{{\small Standard deviation of mean SS scores over the 5 different partitions of Reddit test dataset.}}
\label{tab:Reliability}
\end{table}

\subsection{Response Echo Index}
\begin{figure}[tbp]
\centering
\minipage{.23\textwidth}
\includegraphics[width=\linewidth]{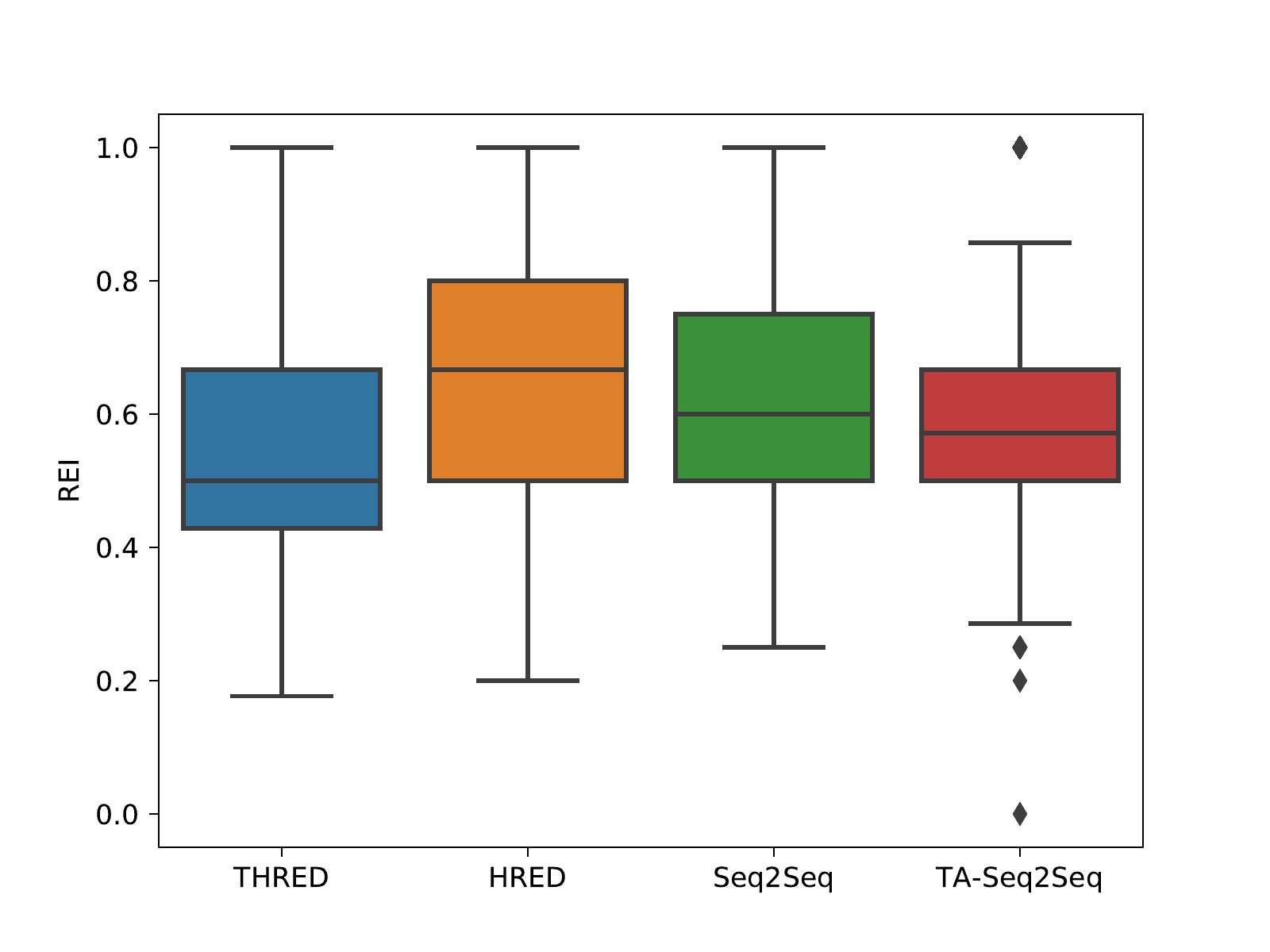}
\subcaption{Reddit}
\label{fig:rei_reddit}
\endminipage \hfill
\minipage{.23\textwidth}
\includegraphics[width=\linewidth]{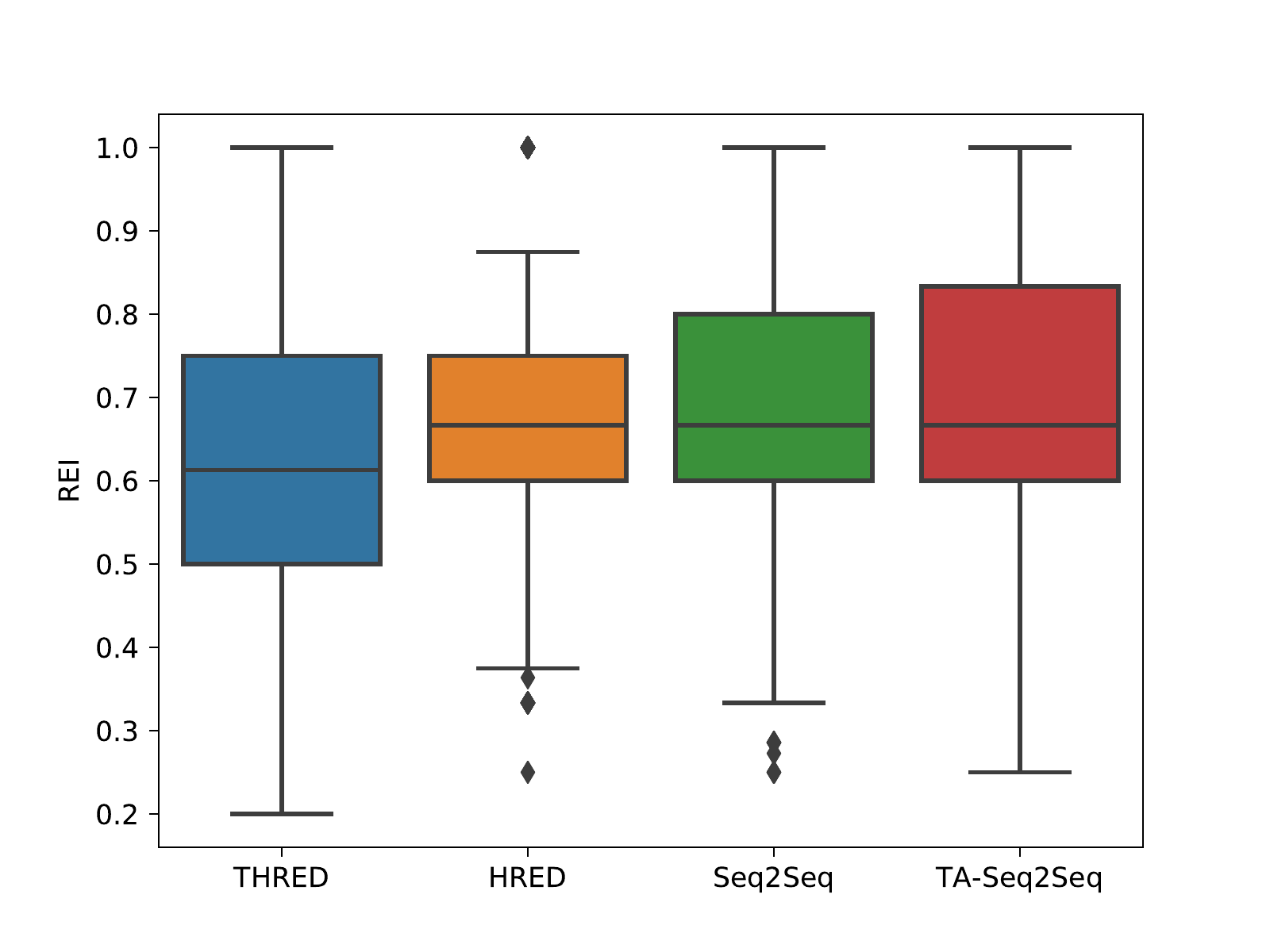}
\subcaption{OpenSubtitles}
\label{fig:rei_OpenSubtitles}
\endminipage
\caption{{\small Performance results of the generated responses from different models based on REI. From left to right, the labels in horizontal axis are THRED, HRED, Seq2Seq, TA-Seq2Seq.}}
\label{fig:REI}
\end{figure}

The goal of the Response Echo Index (REI) metric is to detect overfitting to the training dataset. More specifically, we want to measure the extent to which the responses generated by our model repeat the utterances appearing in the training data. Our approach is close to sampling and finding the nearest neighbour in image generative models \cite{theis2016note}. We randomly sampled 10\% of the training data of both OpenSubtitles and Reddit. The nearest neighbour is determined via Jaccard similarity function.
 Each utterance is represented by lemmatized bag-of-words where stop words and punctuation marks are omitted. In effect, REI is defined as:

\vspace{-1em}
\begin{equation*} \label{eq:REI}
    REI \left(resp_i\right) = \max_{utt_m \in \mathbb{T}_{0.1}} \mathcal{J}(\overline{resp_i}, \overline{utt_m})
\end{equation*}
\vspace{-1em}

where $\bar{t}$ is the normalized form of text $t$, $\mathbb{T}_{0.1}$ denotes the sampled training data, and $\mathcal{J}$ represents Jaccard function.
REI is expected to be low since the generated responses should be distant from the nearest neighbor. According to the results, presented in Figure~\ref{fig:REI}, the REI scores of the responses generated from THRED are the lowest compared to the rest of the models. Such observation leads us to the conclusion that
THRED is able to generate unique responses which appear to be drawn from the input distribution, while being measurably far from the input dataset. This strength in THRED is attributed to the topic attention and incorporating topic words in response generation. Due to the same reason, standard Seq2Seq and HRED fall short.

\subsection{Degree of Diversity \& Perplexity}
To account further for diversity in generated responses, following \cite{li2015diversity}, we calculated $distinct$-$1$ and $distinct$-$2$ by counting unique unigrams and bigrams, normalized by the number of generated words. The results, given in Table~\ref{tab:diversity}, on Reddit
indicate that THRED yields content rich and diverse responses, mainly ascribed to incorporating new topic words into response generation. Further, in perplexity, THRED performs slightly better. 

\subsection{Human Evaluation}
Besides the quantitative measures, 4-scale and side-by-side human evaluation were carried out. Five human raters were recruited for the purpose of evaluating the quality of the responses.  They were fluent, native English speakers and  well-instructed for the judgment task to ensure quality rating. We showed every judge 300 conversations (150 dialogues from Reddit and 150 dialogues from OpenSubtitles) and two generated responses for each dialogue: one generated by THRED model and the other one generated by one of our baselines. The source models were unknown to the evaluators. The responses were ordered in a random way to avoid biasing the judges. Additionally, Fleiss' Kappa score is used to gauge the reliability of the agreement between human evaluators \cite{shao2017generating}. An example of generated responses from the Reddit dataset are provided in Table~\ref{tab:SampleResponsesReddit}
For the 4-scale human evaluation, judges were asked to judge the responses from Bad (0) to Excellent (3). Additional details are provided in the supplementary material.
The results of this experiment, conducted on Reddit, are detailed in Table~\ref{tab:HumanEvaluation_4_side}. The lablers with a high consensus degree rated 32.9\% and 36.9\% of the THRED responses in OpenSubtitles and Reddit respectively as Excellent, which is greatly larger than all baselines (up to 11.6\% and 22.7\% respectively). Apart from the 4-scale rating, we conducted the evaluations side-by-side to measure the gain in THRED over the strong baselines. Specific comparison instructions are included in the supplementary material.
The results, illustrated in Table~\ref{tab:HumanEvaluation_4_side}, suggest that THRED is substantially superior to all baselines in producing informative and plausible responses from human's perspective. The high Kappa scores imply that a major agreement prevails among the lablers. In particular, THRED beats the strong baselines in 52\% of the test data in Reddit (the percentage is achieved by averaging the win ratio). However, for the rest of the cases, THRED is equally good with the baselines in 25\% in Reddit (calculated similarly based on Table \ref{tab:HumanEvaluation_4_side}). Hence, the ratio of cases where THRED is better than or equal with the baselines in terms of quality is 77\% in Reddit.

\begin{table}[t]
\centering
\begin{tabular}{ l | c | c c }
 \hline
  \textbf{ Method} & \textbf{ PPL} & \textbf{ \textit{distinct-1}} & \textbf{ \textit{distinct-2}} \\
 \hline
\small Seq2Seq & \small 62.12 & \small 0.0082 & \small 0.0222 \\
 \small HRED & \small 63.00 & \small 0.0083 & \small 0.0182 \\
 \small TA-Seq2Seq & \small 62.40 & \small 0.0098 & \small 0.0253 \\
 \small THRED & \small \textbf{61.73} & \small \textbf{0.0103} (\textit{+5\%}) & \small \textbf{0.0347} (\textit{+37\%}) \\
 \hline
\end{tabular}
\caption{{\small Performance results of diversity and perplexity metrics of all the models on the Reddit test dataset. THRED surpasses all the baselines with a gain of 5\% in $distinct$-1 and 37\% in $distinct$-2 over TA-Seq2Seq (second best).}}
\label{tab:diversity}
\end{table}

\begin{table*}[t]
\begin{center}
 \begin{tabular}{l|c c c c|c} 
 \hline
  \textbf{Side-by-Side } & \textbf{Wins} & \textbf{Losses} & \textbf{Equally Good} & \textbf{Equally Bad} & \textbf{Kappa} \\
 \hline
 THRED vs Seq2Seq & \textbf{47.5\%{\small $\pm$4.4\%}} & 19.1\%{\small $\pm$3.3\%} & 28.5\%{\small $\pm$3.1\%} & 4.9\%{\small $\pm$1.8\%}  & 0.80\\
 
 THRED vs HRED & \textbf{51.7\%{\small $\pm$4.6\%}} & 20.1\%{\small $\pm$3.4\%} & 20.9\%{\small $\pm$3.1\%} & 7.2\%{\small $\pm$2.3\%} & 0.75 \\
 
 THRED vs TA-Seq2Seq & 
 \textbf{55.7\%{\small $\pm$4.1\%}} & 13.5\%{\small $\pm$2.6\%} & 24.7\%{\small $\pm$3.0\%} & 6.1\%{\small $\pm$1.8\%} & 0.77\\
  \hline
  \textbf{4-scale } & \textbf{Excellent} & \textbf{Good} & \textbf{Poor} & \textbf{Bad} & \textbf{Kappa} \\
  \hline
  Seq2Seq & 22.7\%{\small $\pm$2.6\%} & 47.2\%{\small $\pm$3.5\%} & 22.5\%{\small $\pm$3.5\%} & 7.6\%{\small $\pm$2.7\%}  & 0.80\\
 
 HRED & 14.5\%{\small $\pm$2.8\%} & 46.7\%{\small $\pm$3.8\%} & 31.3\%{\small $\pm$3.8\%} & 7.5\%{\small $\pm$2.5\%} & 0.84 \\
 
 TA-Seq2Seq & 
 17.1\%{\small $\pm$2.4\%} & 44.8\%{\small $\pm$3.5\%} & 30.1\%{\small $\pm$3.2\%} & 8.0\%{\small $\pm$2.3\%} & 0.72\\
 
 THRED & 
 \textbf{36.9\%{\small $\pm$3.0\%}} & 51.1\%{\small $\pm$2.9\%} & 10.3\%{\small $\pm$2.4\%} & 1.7\%{\small $\pm$1.5\%} & 0.84\\
  \hline
\end{tabular}
\caption{Side-by-side human evaluation along with 4-scale human evaluation of dialogue utterance prediction on Reddit dataset (mean preferences {\small $\pm$90\%} confidence intervals).}
\label{tab:HumanEvaluation_4_side}
\end{center}
\end{table*}

\begin{figure}[!tb]
\centering
\minipage{.24\textwidth}
\includegraphics[width=\linewidth]{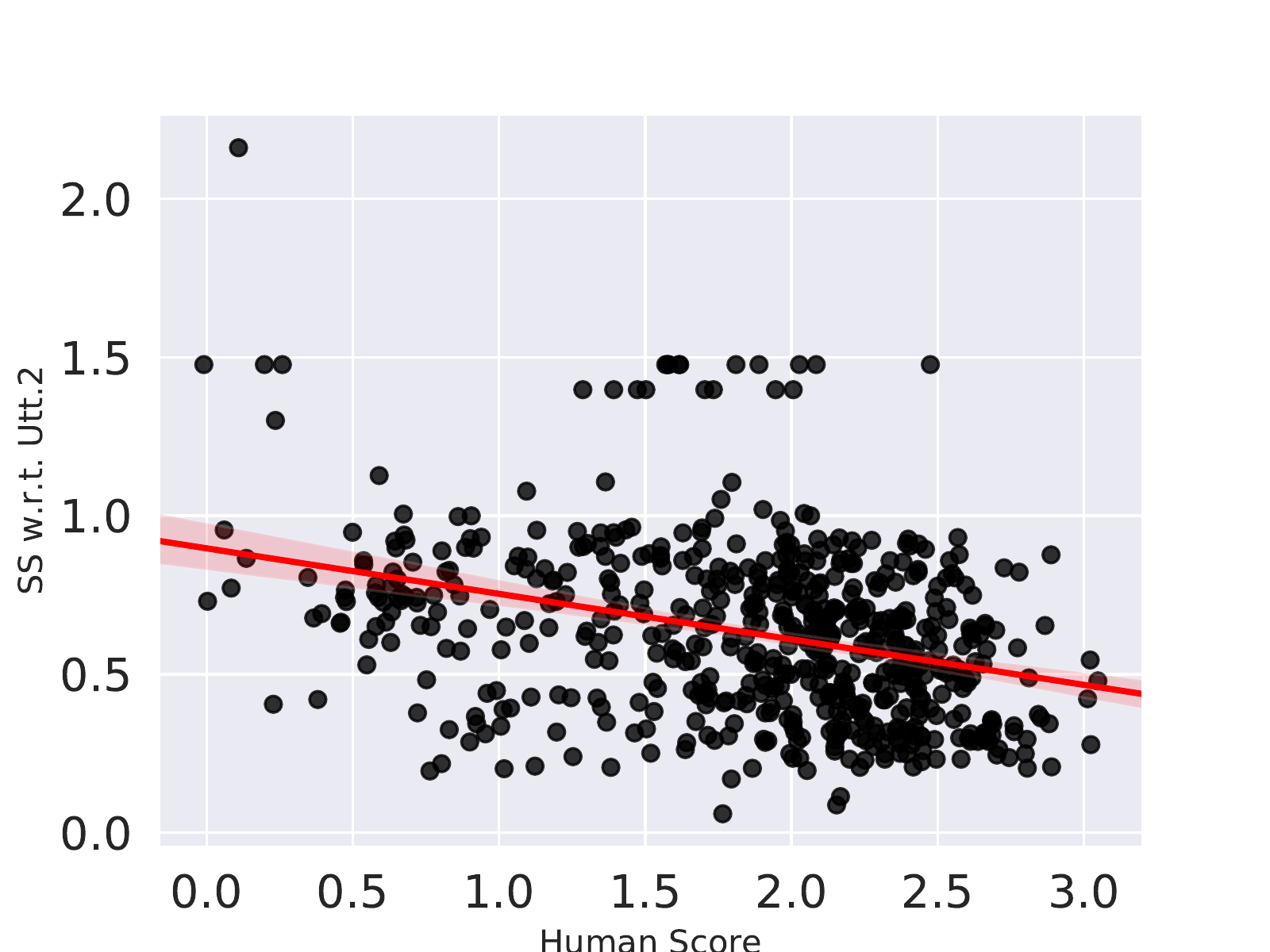}
\subcaption{{\scriptsize SS w.r.t. Utt.2 ($\rho=-0.341$)}}
\label{fig:corr_reddit_ss_utt2}
\endminipage
\minipage{.24\textwidth}
\includegraphics[width=\linewidth]{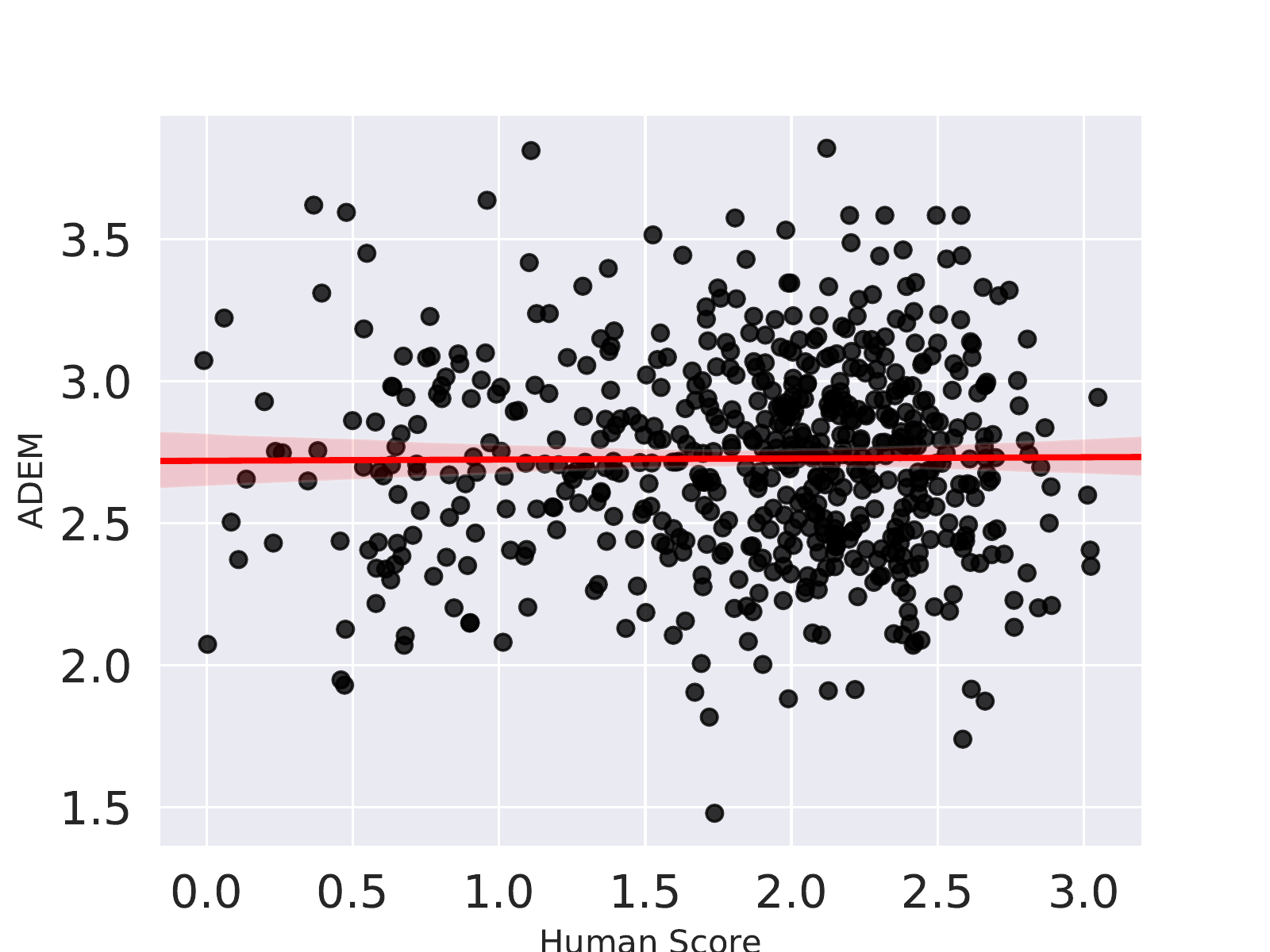}
\subcaption{{\scriptsize ADEM ($\rho=0.014$)}}
\label{fig:corr_reddit_adem}
\endminipage
\caption{{\small Scatter plots illustrating correlation between automated metrics and human judgment (Pearson correlation coefficient is reported in the brackets). In order to better visualize the density of the points, we added stochastic noise generated by Gaussian distribution $\mathcal{N}(0, 0.1)$ to the human ratings (i.e., horizontal axis) at the cost of lowering correlation, as done in \cite{lowe2017towards}.}%
}
\label{fig:correlation_scatter}
\end{figure}

\subsubsection{Automated metric vs. Human evaluation} 
We also carried out an analysis on the correlation between the human evaluator ratings and our quantitative scores. The Semantic Similarity metric, which requires no pre-training, reaches a Pearson correlation of -0.341 with respect to the most recent utterance (Utt.2) on Reddit.  A negative correlation is anticipated here since the higher human ratings correspond to the lower semantic distance. This compares with values of 0.351 for Automatic User Ratings \cite{venkatesh2018evaluating} and 0.436 for ADEM \cite{lowe2017towards} from recent models which required large amounts of training data and computation.
The correlations are visualized as scatter plots in Figure~\ref{fig:correlation_scatter}. In addition, we assessed ADEM on our test datasets using the pre-trained weights{\footnote{\url{https://github.com/mike-n-7/ADEM}}}, provided by the authors. ADEM achieves low correlation with human judgment ($\rho=0.014$ on Reddit and $\rho=0.034$ on OpenSubtitles) presumably since the quality of its predicted scores highly depends on the corpus on which the model is trained.

\subsection{Comparing Datasets}
Finally, we investigate the impact of training datasets on the quality of the responses generated by THRED and all baselines. Table \ref{tab:dataset_comparison} has results which support that our cleaner, well-parsed Reddit dataset generates significantly improved responses over our metrics of interest.  In particular, 
we contrast the two datasets in terms of human judgment and the automated metrics among all the models. Regarding human assessment, we took the mean evaluation rating (MER) per response in the test data to draw the comparison between the datasets. As demonstrated in Table~\ref{tab:dataset_comparison} (see more details in Figure~\ref{fig:dataset_comparison} in the Appendix), the human evaluators scored generated responses from the Reddit dataset higher than utterances generated from the OpenSubtitles dataset, which is true not only in THRED, but in all models. Consequently, the training data plays a crucial role in generating high-quality responses. Morever, in OpenSubtitles, the assumption of spotting a conversation, as stated in Section~\ref{sec:datasets}, tends to include extraneous utterances in the dialogue, impeding  the response generation process. While such presumption may seem valid in dealing with two-turn dialogues, it can aggravate the quality of conversations in multi-turn dialogues.

\begin{table}[t]
\centering
\begin{tabular}{ l | c c | c c}
 \hline
  \multirow{2}{*}{\small \textbf{Method}} & \multicolumn{2}{c|}{{\small \textbf{OpenSubtitles}}} & \multicolumn{2}{c}{{\small \textbf{Reddit}}} \\
  & $\mu$ & $\sigma$ & $\mu$ & $\sigma$ \\
 \hline
 {\small Human MER} & {\small 1.681} & {\small 0.639} & {\small \textbf{1.868}} & {\small 0.624}\\
 {\small SS w.r.t. Utt.1} & {\small 0.642} & {\small 0.167} & {\small \textbf{0.631}} & {\small 0.270} \\
 {\small SS w.r.t. Utt.2} & {\small 0.662} & {\small 0.209} & {\small \textbf{0.599**}} & {\small 0.262} \\
 {\small REI} & {\small 0.667} & {\small 0.205} & {\small \textbf{0.546**}} & {\small 0.201}\\
 \hline
\end{tabular}
\caption{{\small Mean $\mu$ and standard deviation $\sigma$ over metrics per dataset to fare Reddit against OpenSubtitles. (\textbf{**} indicates statistical significance with $p$-value $<$ 0.001)
}}
\label{tab:dataset_comparison}
\end{table}

\section{Conclusion}

In this work, we introduce the Topical Hierarchical Recurrent Encoder Decoder (THRED) model for generating topically consistent responses in multi-turn open conversations.
We demonstrate that THRED significantly outperforms current state-of-the-art systems on quantitative metrics and human judgment.
Additionally, we evaluate our new model and existing models with two new metrics which prove to be good measures for automatically evaluating the quality of the responses.
Finally, we present a parsed and cleaned dataset based on conversations from Reddit which improves generated responses. 
We expect more advanced work to be done in the area of chit-chat dialogue to improve the models, training data, and means of evaluation.

\bibliography{acl2019}
\bibliographystyle{acl_natbib}

\clearpage
\appendix
\section{Supplementary Material}

\subsection{Experimental Setup}
The model parameters are learned by optimizing the log-likelihood of the utterances via Adam optimizer with a learning rate of 0.0002; we followed \cite{luong2015effective} for decaying the learning rate.
The dropout rate is set to 0.2 for both the encoder and the decoder to avoid overfitting.
For all the baselines, we experimented hidden state units with the size of 1024. For our model, we tested with encoder and decoder hidden state units of size 800, the same for the context encoder. 
During inference, we experimented with the standard beam search with the beam width 5 and the length normalization $\alpha =1$ \cite{wu2016google}. We noticed that applying the length normalization resulted in a more diverse and longer sentences but at the expense of the semantic coherence of the response in some cases.

\textbf{Training LDA model}: We trained two LDA models\footnote{We used LDA model developed in Gensim library.}:
one trained on OpenSubtitles and the other one trained on Reddit. Both of them were trained on 1M dialogues. We set the number of topics to 150, $\alpha$ to $\frac{1}{150}$ and $\gamma$ to 0.01.
We filtered out stop words and universal words. We also discarded the 1000 words with the highest frequency from the topic words.

\subsection{Human Evaluation Procedure}

For the 4-scale human evaluation, judges were asked to judge the responses
from Bad (0) to Excellent (3). Excellent (score 3): The response is very appropriate,
on topic, fluent, interesting and shows understanding of the context.
Good (score 2): The response is coherent with the context but it is not diverse
and informative. It may imply the answer. Poor (score 1): The response is interpretable
and grammatically correct but completely off-topic. Bad (score 0):
The response is grammatically broken and it does not provide an answer.
Regarding the side-by-side evaluation, humans were asked
to favor response 1 over response 2 if: (1) response 1 is relevant, logically
consistent to the context, fluent and on topic; or (2) Both responses 1 and 2 are relevant, consistent and fluent but response 1 is more informative than
response 2. If judges cannot tell which one is better, they can rate the responses
as “Equally good” or “Equally Bad”.

\begin{figure*}[!tb]
\centering
\minipage{.48\textwidth}
\includegraphics[width=\linewidth]{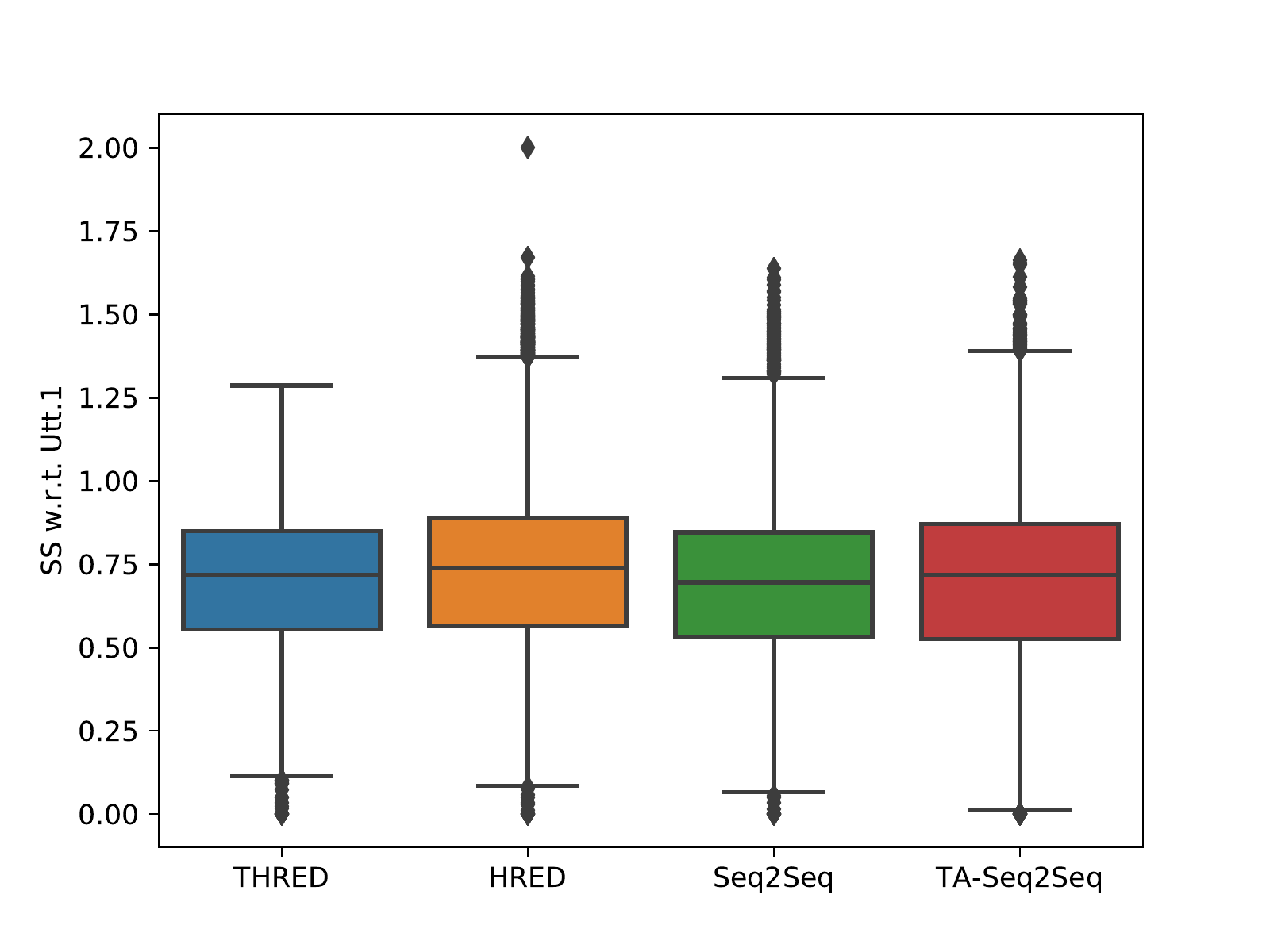}
\subcaption{{\scriptsize SS w.r.t. Utt.1 on Reddit (0.678, 0.722, 0.682, 0.686)}}
\label{fig:semantic_reddit_utt1}
\endminipage \hfill
\minipage{.48\textwidth}
\includegraphics[width=\linewidth]{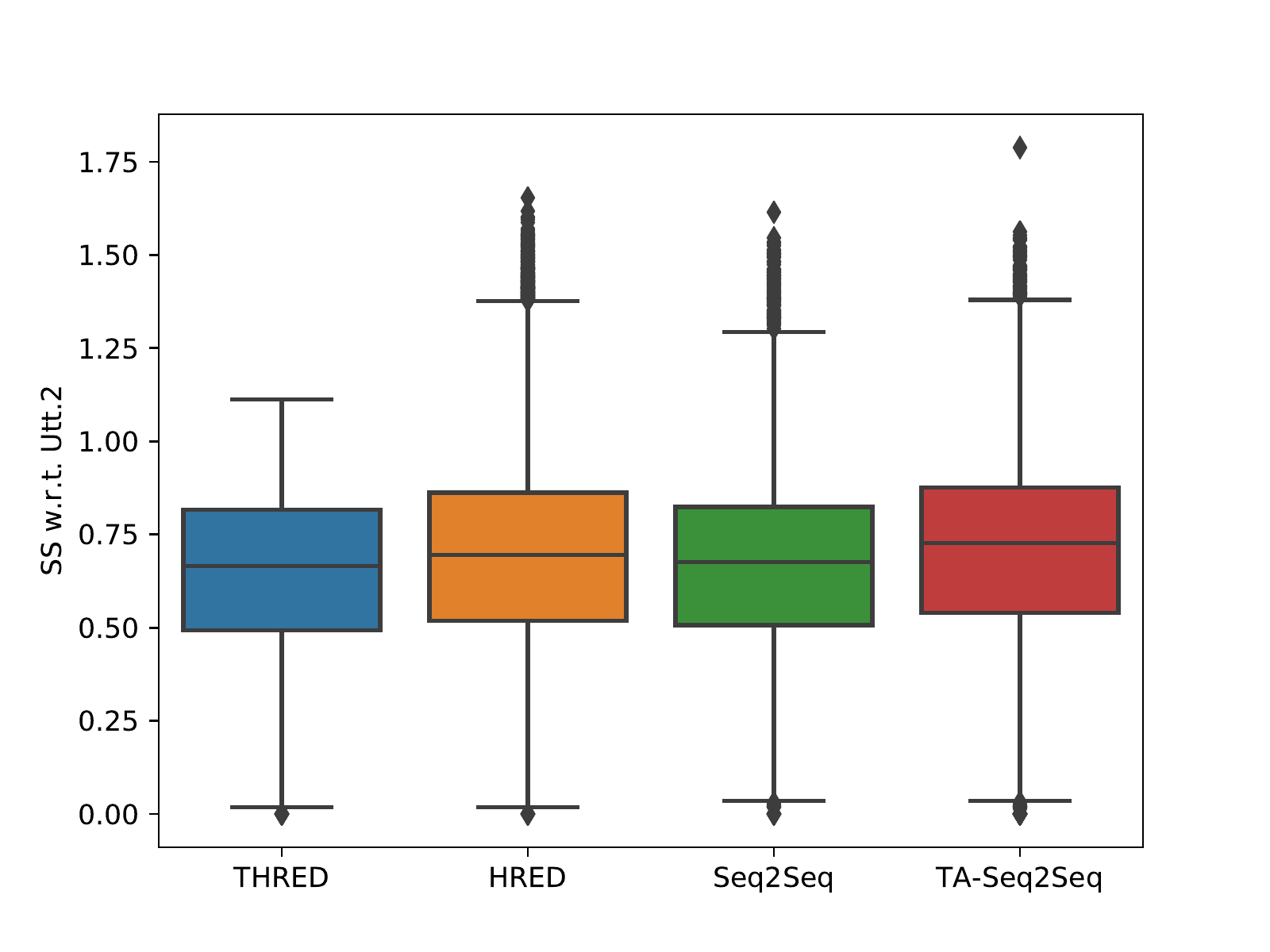}
\subcaption{{\scriptsize SS  w.r.t. Utt.2 on Reddit (0.631**, 0.679, 0.656, 0.697)}}
\label{fig:semantic_reddit_utt2}
\endminipage \hfill
\minipage{.48\textwidth}
\includegraphics[width=\linewidth]{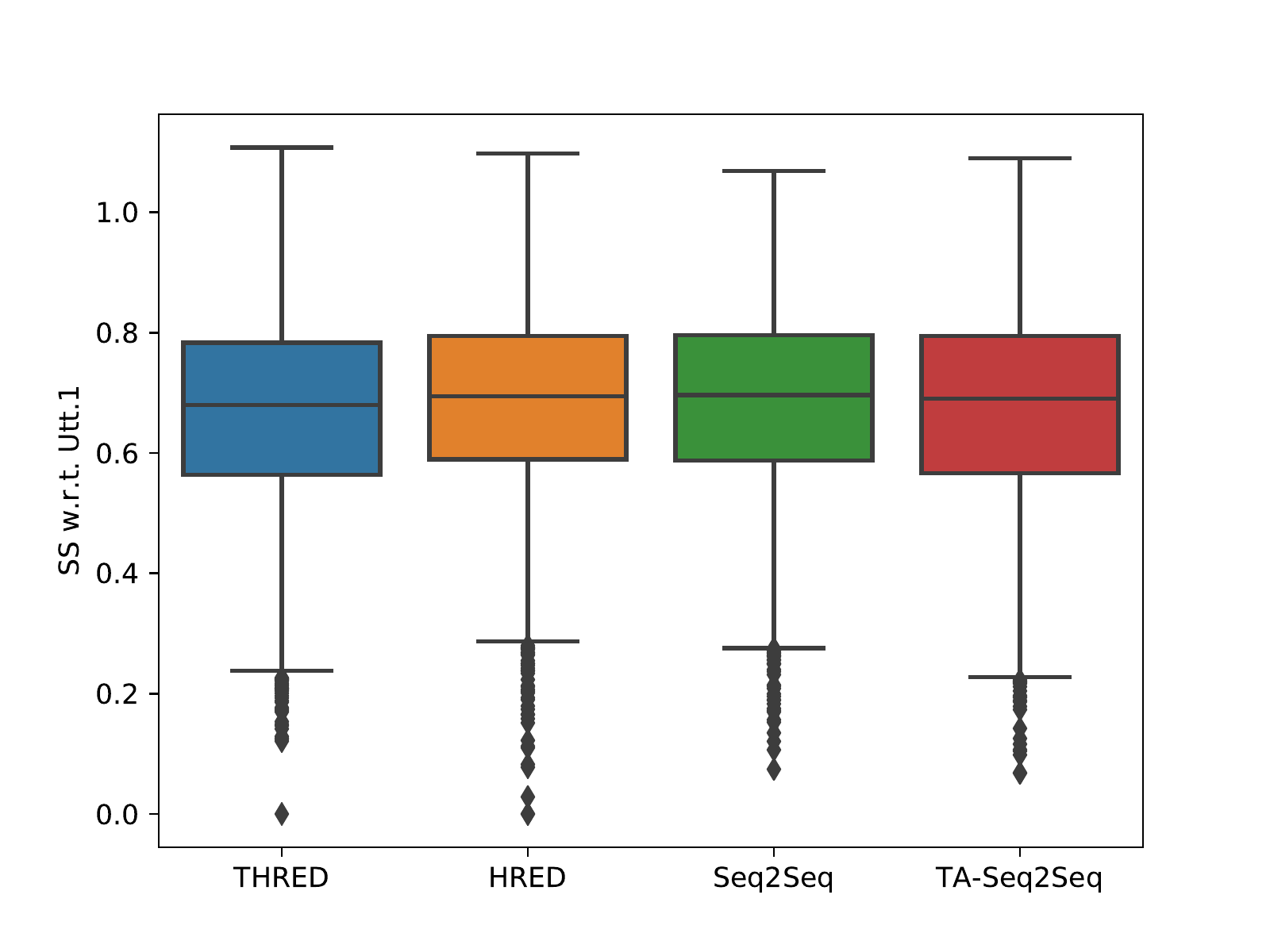}
\subcaption{{\scriptsize SS w.r.t. Utt.1 on OpenSubtitles (0.679, 0.694, 0.696, 0.690)}}
\label{fig:semantic_opensubtitles_utt1}
\endminipage \hfill
\minipage{.48\textwidth}
\includegraphics[width=\linewidth]{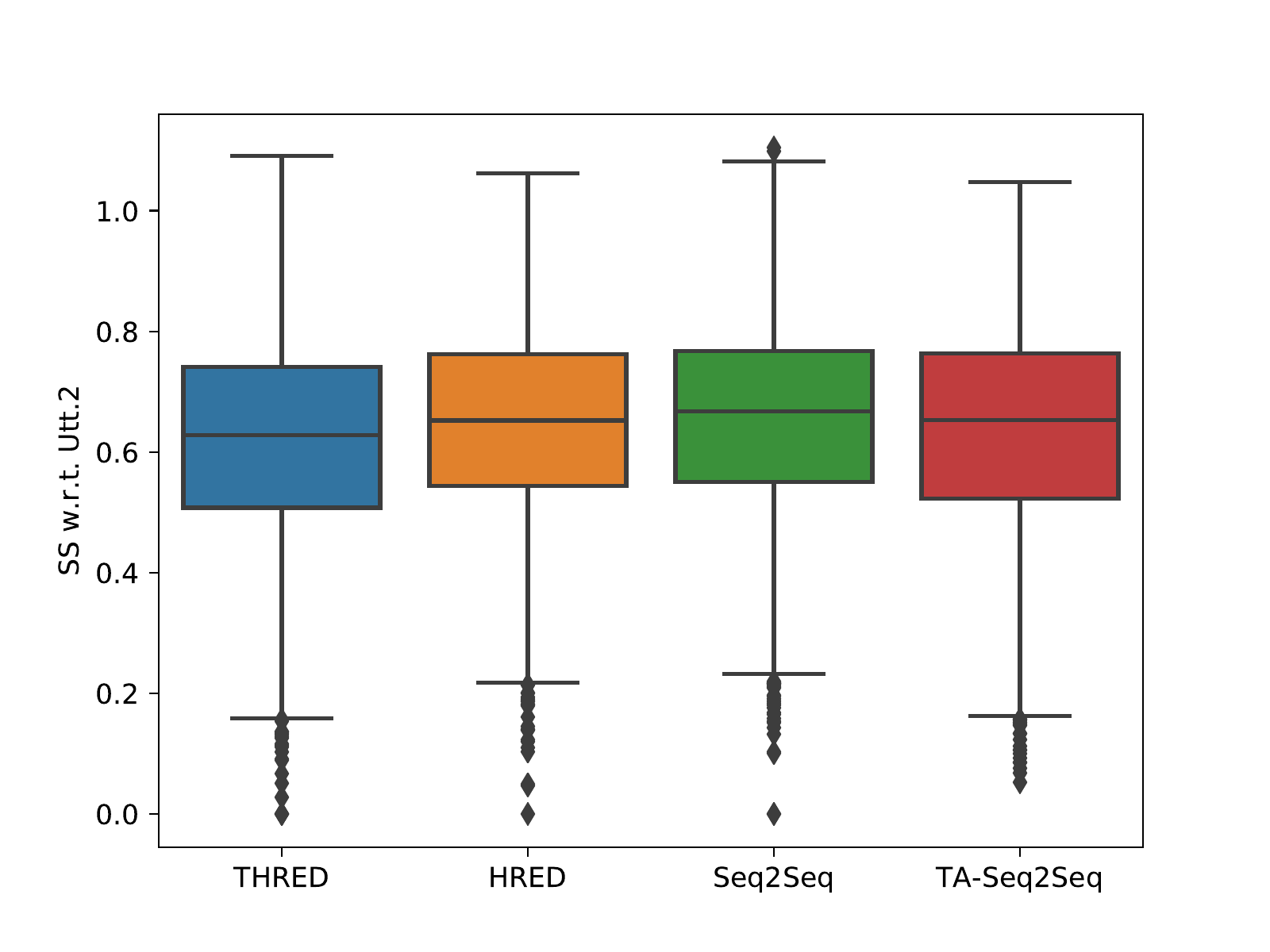}
\subcaption{{\scriptsize SS w.r.t. Utt.2 on OpenSubtitles (0.628**, 0.653, 0.667, 0.653)}}
\label{fig:semantic_opensubtitles_utt2}
\endminipage
\caption{{\small Box plots showcasing the performance of the generated responses from different models based on the Semantic Similarity metric with respect to Utt.1 and Utt.2 (complementary to Table \ref{tab:SC}). From left to right, the labels in horizontal axis are THRED, HRED, Seq2Seq and TA-Seq2Seq.
THRED surpasses all baselines in similarity with Utt.2, and works mildly better in similarity with Utt.1. The median for every box plot is also reported between brackets. The order of the values is the same as the order of the methods.}}
\label{fig:semantic_similarity}
\end{figure*}

\begin{table*}[t]
\linespread{0.8}\selectfont\centering
\centering
\begin{tabular}{l|p{9.5cm}} 
 	\hline
    {\small \textsc{Context}} & {\small \textsc{Generated Responses}} \\
    \hline
       
        \multirow{4}{6.5cm}{{\footnotesize \textbf{(OpenSubtitles)} the giant observatory in hawaii are like time machines capturing the faint light that has travelled towards us through all of cosmic history \textcolor{blue}{\textbf{$\rightarrow$}} the deeper they look into space the farther back they see in time}}
        & {\footnotesize \textbf{THRED:} as the \textcolor{red}{stars} move into \textcolor{red}{space} they see the rising \textcolor{red}{sun} and the stars rise above the \textcolor{red}{clouds} (\underline{{\em Excellent, Excellent, Good, Excellent, Poor}})} \\
        & {\footnotesize \textbf{HRED:} 
this is the first time we have ever seen (\underline{{\em Poor, Poor, Bad, Good, Bad}})} \\
        & {\footnotesize \textbf{Seq2Seq:}the stars and stars of the moon are the stars and stars of the stars and stars and stars and stars and stars and stars (\underline{{\em Poor, Bad, Bad, Bad, Poor}})} \\
        & {\footnotesize \textbf{TA-Seq2Seq:} 
they have been searching for a long time for a long time (\underline{{\em Good, Poor, Good, Good, Good}})} \\
    \hline
        \multirow{4}{6.5cm}{{\footnotesize \textbf{(OpenSubtitles)} and closer to home in baltimore prominent psychiatrist and the author of a newly released book on insanity disappeared mysteriously last night after a lecture at the university \textcolor{blue}{$\rightarrow$}  former mental patient james cole is wanted for questioning regarding dr cole 's disappearance}}
        & {\footnotesize \textbf{THRED:} \textcolor{red}{authorities} warn that cole has a history of \textcolor{red}{violence} and advise anyone to notify authorities (\underline{{\em Excellent, Excellent, Excellent, Excellent, Excellent}})}  \\
        & {\footnotesize \textbf{HRED:} they have been unable to identify him (\underline{{\em Good, Poor, Poor, Good, Poor}}) } \\
        & {\footnotesize \textbf{Seq2Seq:}cole cole is cole is cole(\underline{{\em Poor, Bad, Bad, Bad, Bad}})} \\
        & {\footnotesize \textbf{TA-Seq2Seq:} \textcolor{red}{authorities} warn that cole has been \textcolor{red}{kidnapped} (\underline{{\em Poor, Good, Good, Good, Good}})} \\
    \hline
\end{tabular}
\linespread{1.0}
\caption{One cherry-picked dialogues out of 150 conversations along with the generated responses from all models. Human judgments are provided in the brackets. 
The blue arrow specifies a dialogue exchange and the highlighted words in red represent the topic words acquired from the pre-trained LDA model. }
\label{tab:SampleResponses}
\end{table*}

\begin{table*}[t]
\centering
\begin{tabular}{ l | c c c | c c c }
 \hline
 \multirow{2}{*}{\textbf{Method}} & \multicolumn{3}{c|}{\textbf{OpenSubtitles}} & \multicolumn{3}{c}{\textbf{Reddit}} \\
  & \textbf{PPL} & \textbf{ \textit{distinct-1}} & \textbf{ \textit{distinct-2}} & \textbf{PPL} & \textbf{\textit{distinct-1}} & \textbf{ \textit{distinct-2}} \\
 \hline
Seq2Seq & 74.37 & 0.0112 & 0.0258 & 62.12 & 0.0082 & 0.0222 \\
HRED & 74.65 & 0.0079 & 0.0219 & 63.00 & 0.0083 & 0.0182 \\
TA-Seq2Seq & 75.92 & 0.0121 & 0.0290 & 62.40 & 0.0098 & 0.0253 \\
THRED & \textbf{73.61} & \textbf{0.0157} (\textit{+30\%}) & \textbf{0.0422} (\textit{+45\%}) & \textbf{61.73} & \textbf{0.0103} (\textit{+5\%}) & \textbf{0.0347} (\textit{+37\%}) \\
 \hline
\end{tabular}
\caption{{ Complete performance results of diversity and perplexity on Reddit test data and OpenSubtitles test data (complementary to Table~\ref{tab:diversity}). The numbers in the bracket indicate the gain of $distinct$-1 and $distinct$-2 over the second best method (i.e., TA-Seq2Seq). }}
\label{tab:diversity_full}
\end{table*}

\begin{table*}[t]
\begin{center}
 \begin{tabular}{l|c c c c|c} 
 \hline
  \textbf{Side-by-Side} & \textbf{Wins} & \textbf{Losses} & \textbf{Equally Good} & \textbf{Equally Bad} & \textbf{Kappa} \\
 \hline

  THRED vs Seq2Seq & \textbf{54.0\%{\small $\pm$4.2\%}} & 18.4\%{\small $\pm$3.4\%} & 17.2\%{\small $\pm$3.0\%} & 10.4\%{\small $\pm$2.3\%} & 0.75 \\
 
 THRED vs HRED & \textbf{51.6\%{\small $\pm$4.4\%}} & 19.5\%{\small $\pm$3.5\%} & 18.4\%{\small $\pm$2.9\%} & 10.5\%{\small $\pm$2.4\%} & 0.72 \\
 
 THRED vs TA-Seq2Seq & 
 \textbf{64.0\%{\small $\pm$4.3\%}} & 14.4\%{\small $\pm$3.1\%} & 14.1\%{\small $\pm$2.5\%} & 7.5\%{\small $\pm$2.1\%} & 0.90 \\
  \hline
  \textbf{4-scale Rating} & \textbf{Excellent} & \textbf{Good} & \textbf{Poor} & \textbf{Bad} & \textbf{Kappa} \\
 \hline
  Seq2Seq & \textbf{8.4\%{\small $\pm$2.2\%}} & 48.9\%{\small $\pm$3.9\%} & 33.2\%{\small $\pm$3.7\%} & 9.5\%{\small $\pm$3.1\%} & 0.89 \\
 
 HRED & \textbf{11.6\%{\small $\pm$2.4\%}} & 41.5\%{\small $\pm$3.4\%} & 36.9\%{\small $\pm$3.9\%} & 10.0\%{\small $\pm$2.8\%} & 0.79 \\
 
 TA-Seq2Seq & 
 \textbf{9.5\%{\small $\pm$2.1\%}} & 42.3\%{\small $\pm$3.7\%} & 34.7\%{\small $\pm$3.9\%} & 13.6\%{\small $\pm$3.7\%} & 0.92 \\
 
 THRED & 
 \textbf{32.9\%{\small $\pm$3.6\%}} & 49.2\%{\small $\pm$3.3\%} & 16.8\%{\small $\pm$3.0\%} & 1.1\%{\small $\pm$0.9\%} & 0.83\\
  \hline
\end{tabular}
\caption{Side-by-side human evaluation along with 4-scale human evaluation of dialogue utterance prediction on OpenSubtitles dataset (mean preferences {\small $\pm$90\%} confidence intervals). Results on Reddit dataset are reported in Table \ref{tab:HumanEvaluation_4_side}.  }
\label{tab:SideBySide_open}
\end{center}
\end{table*}

\begin{figure*}[!tb]
\centering
\minipage{.29\textwidth}
\includegraphics[width=\linewidth]{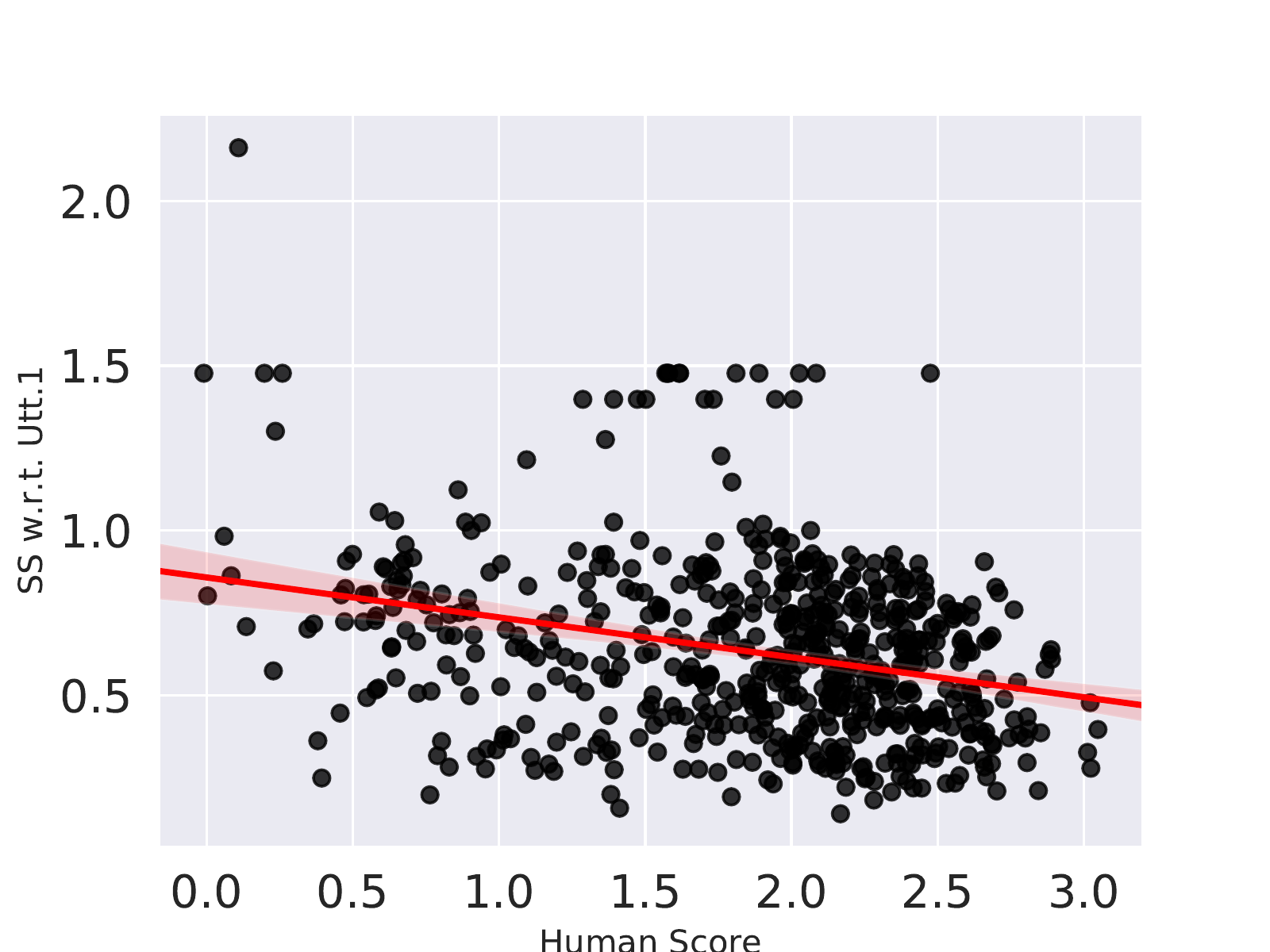}
\subcaption{{\scriptsize \textbf{Reddit:} SS Utt.1 ($\rho=-0.286$)}}
\label{fig:corr_reddit_utt1}
\endminipage
\minipage{.29\textwidth}
\includegraphics[width=\linewidth]{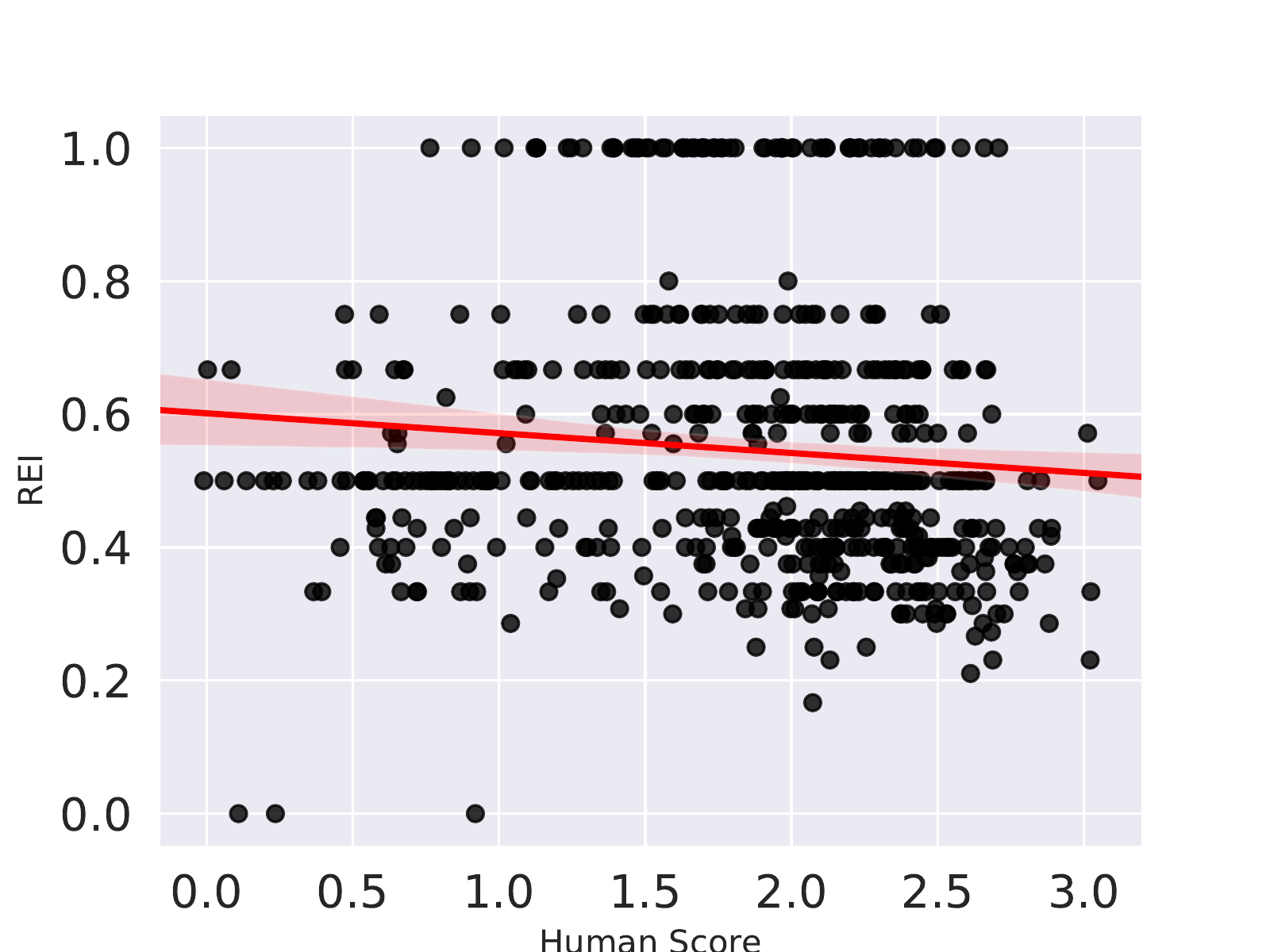}
\subcaption{{\scriptsize \textbf{Reddit:} REI ($\rho=-0.196$)}}
\label{fig:corr_reddit_rei}
\endminipage
\minipage{.29\textwidth}
\includegraphics[width=\linewidth]{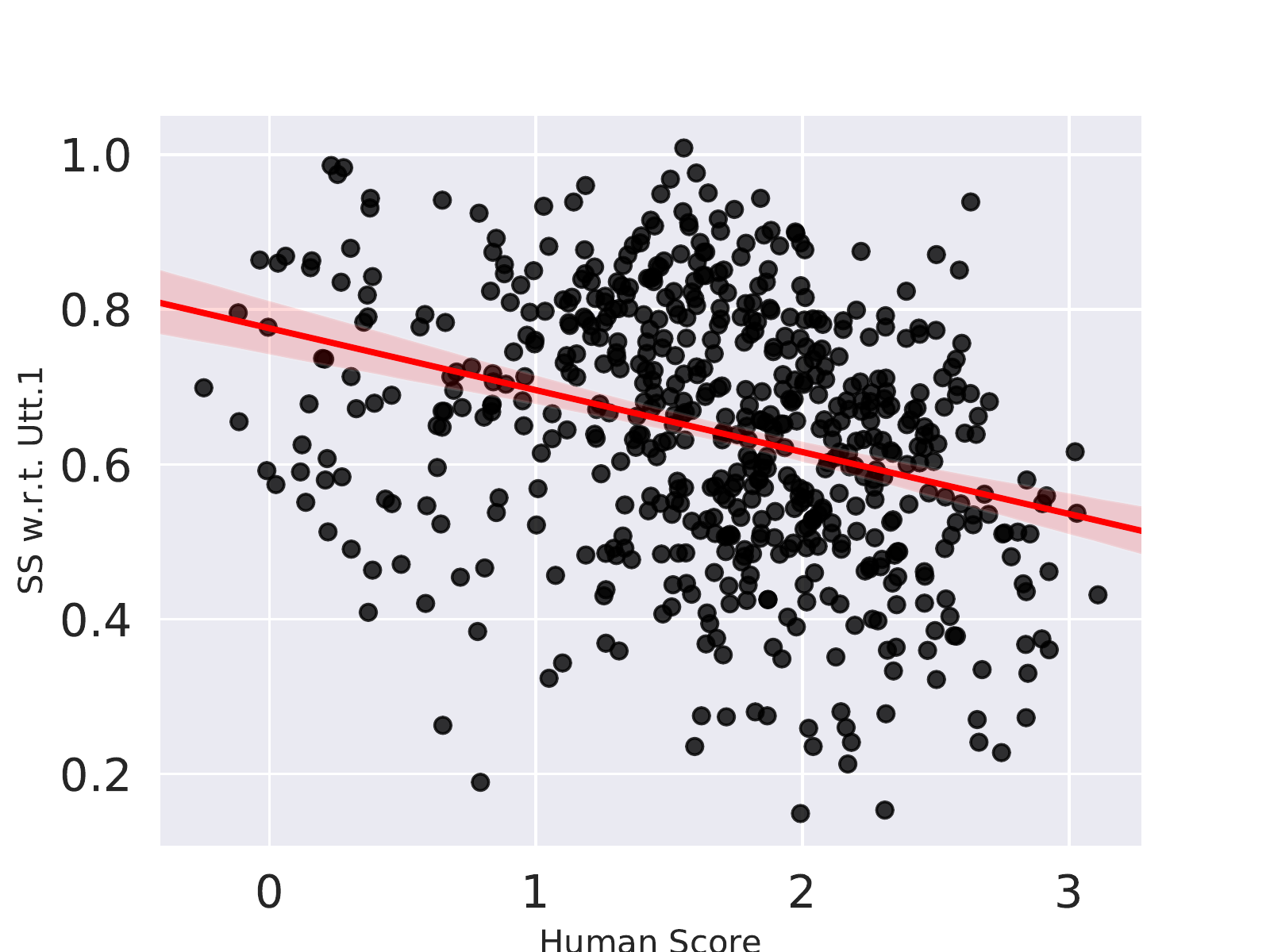}
\subcaption{{\scriptsize \textbf{OpenSubtitles:} SS Utt.1 ($\rho=-0.317$)}}
\label{fig:corr_opensubtitles_utt1}
\endminipage \hfill
\minipage{.29\textwidth}
\includegraphics[width=\linewidth]{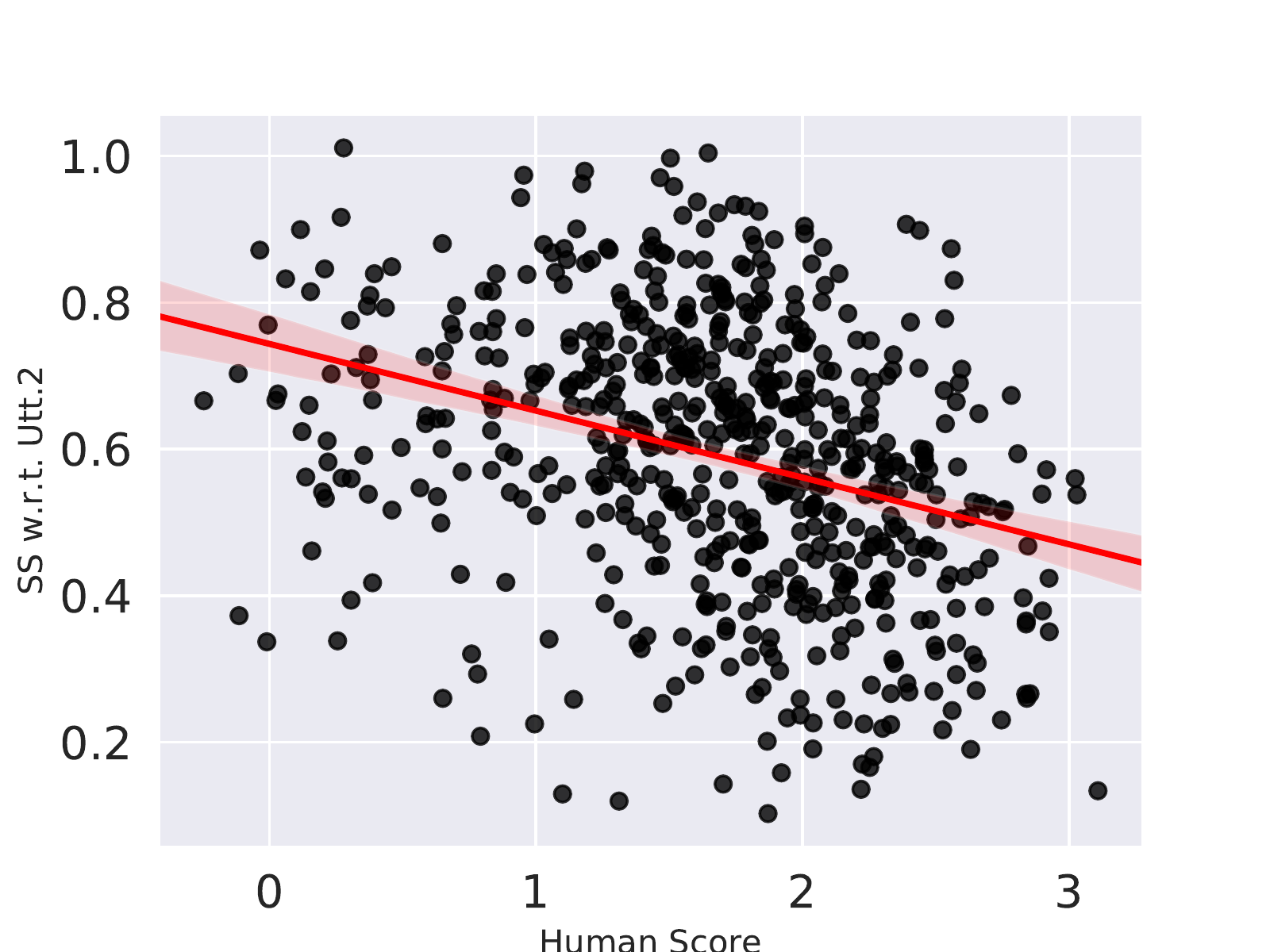}
\subcaption{{\scriptsize \textbf{OpenSubtitles:} SS Utt.2 ($\rho=-0.324$)}}
\label{fig:corr_opensubtitles_utt2}
\endminipage
\minipage{.29\textwidth}
\includegraphics[width=\linewidth]{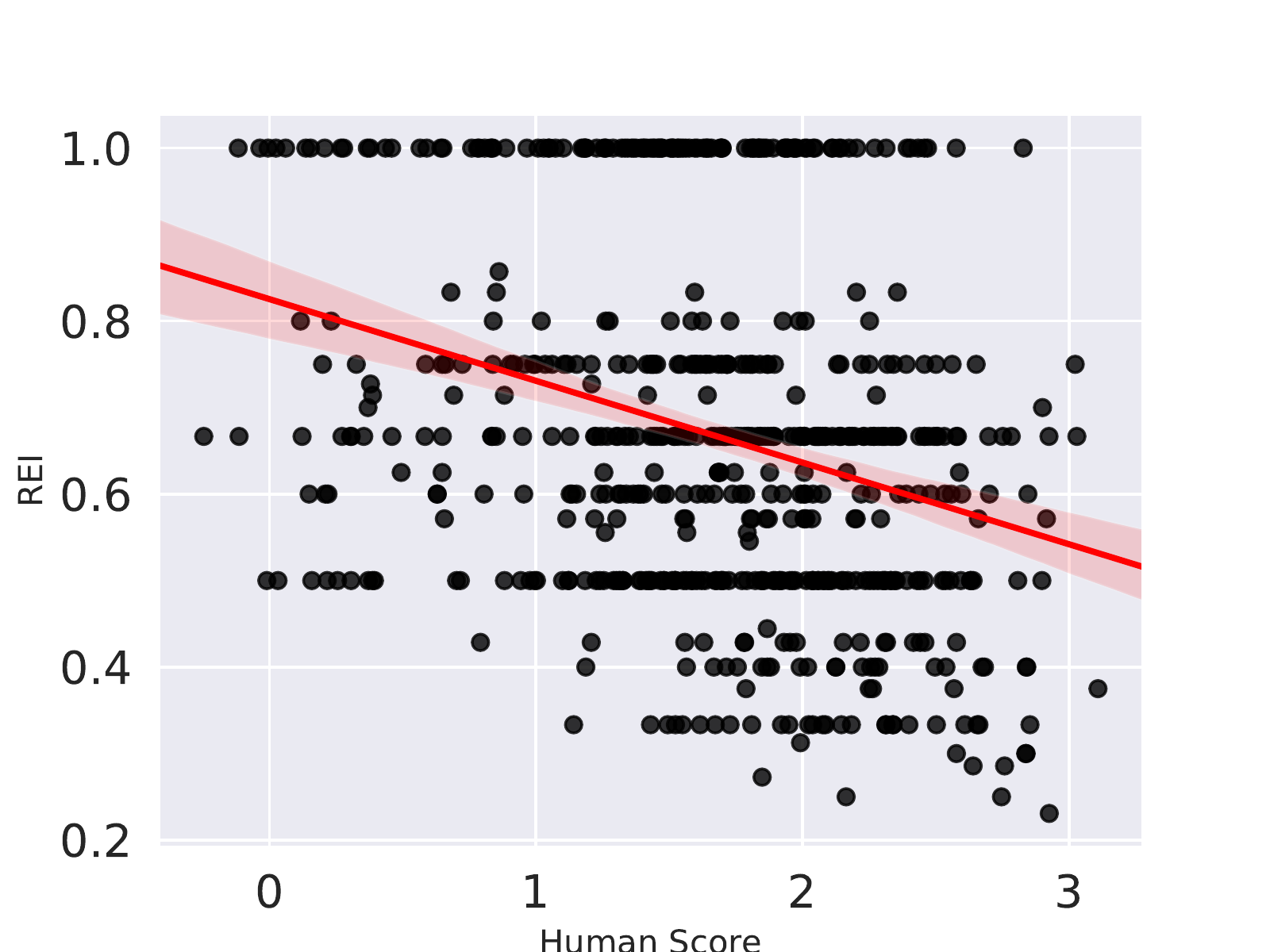}
\subcaption{{\scriptsize \textbf{OpenSubtitles:} REI ($\rho=-0.295$)}}
\label{fig:corr_opensubtitles_rei}
\endminipage
\minipage{.29\textwidth}
\includegraphics[width=\linewidth]{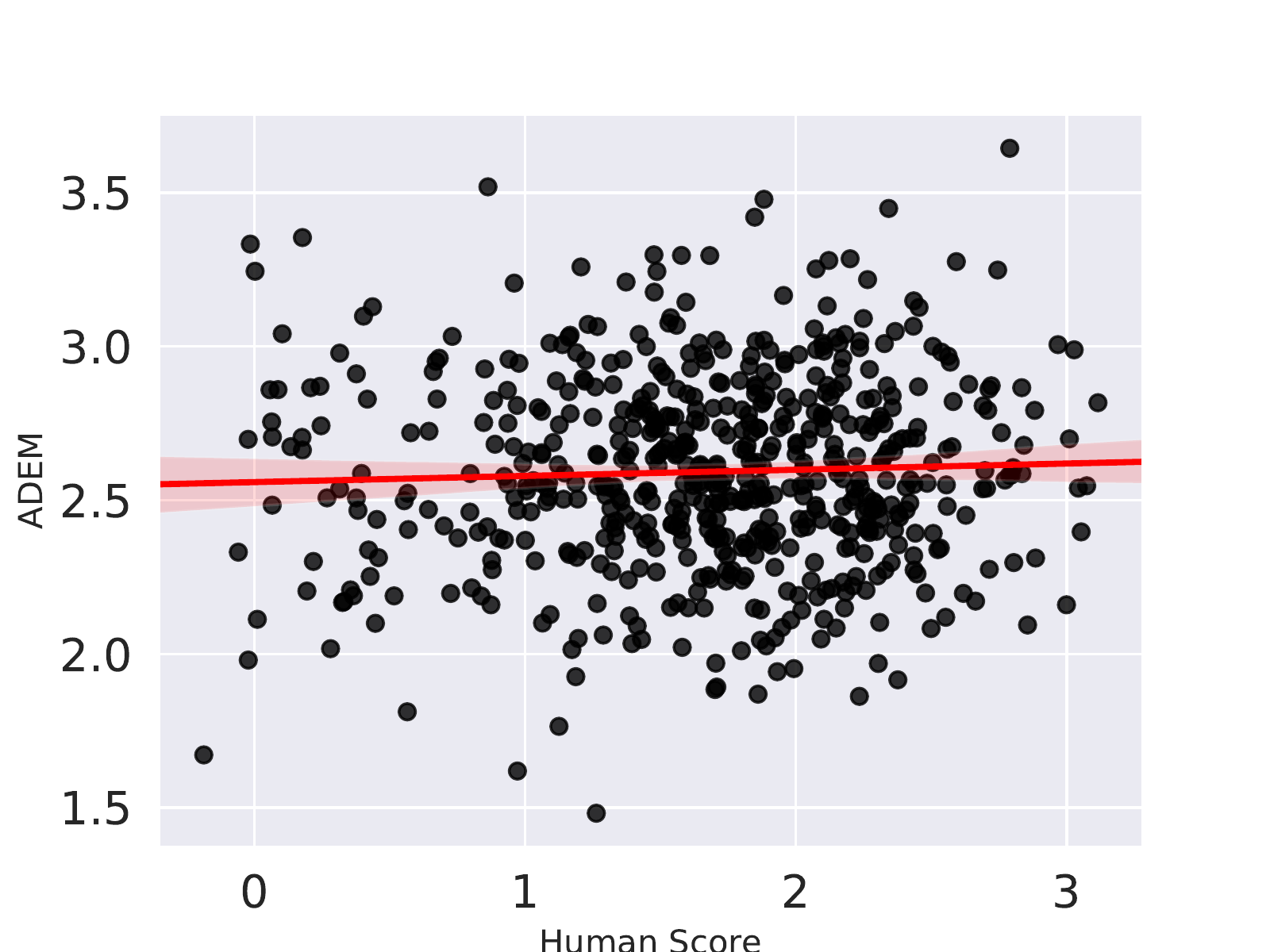}
\subcaption{{\scriptsize \textbf{OpenSubtitles:} ADEM ($\rho=0.034$)}}
\label{fig:corr_reddit_adem}
\endminipage
\caption{Scatter plots illustrating correlation between automated metrics and human judgment (Pearson correlation coefficient is reported in the brackets).%
}
\label{fig:correlation_scatter_full}
\end{figure*}

\begin{figure*}[!tb]
\centering
\minipage{.45\textwidth}
\includegraphics[width=\linewidth]{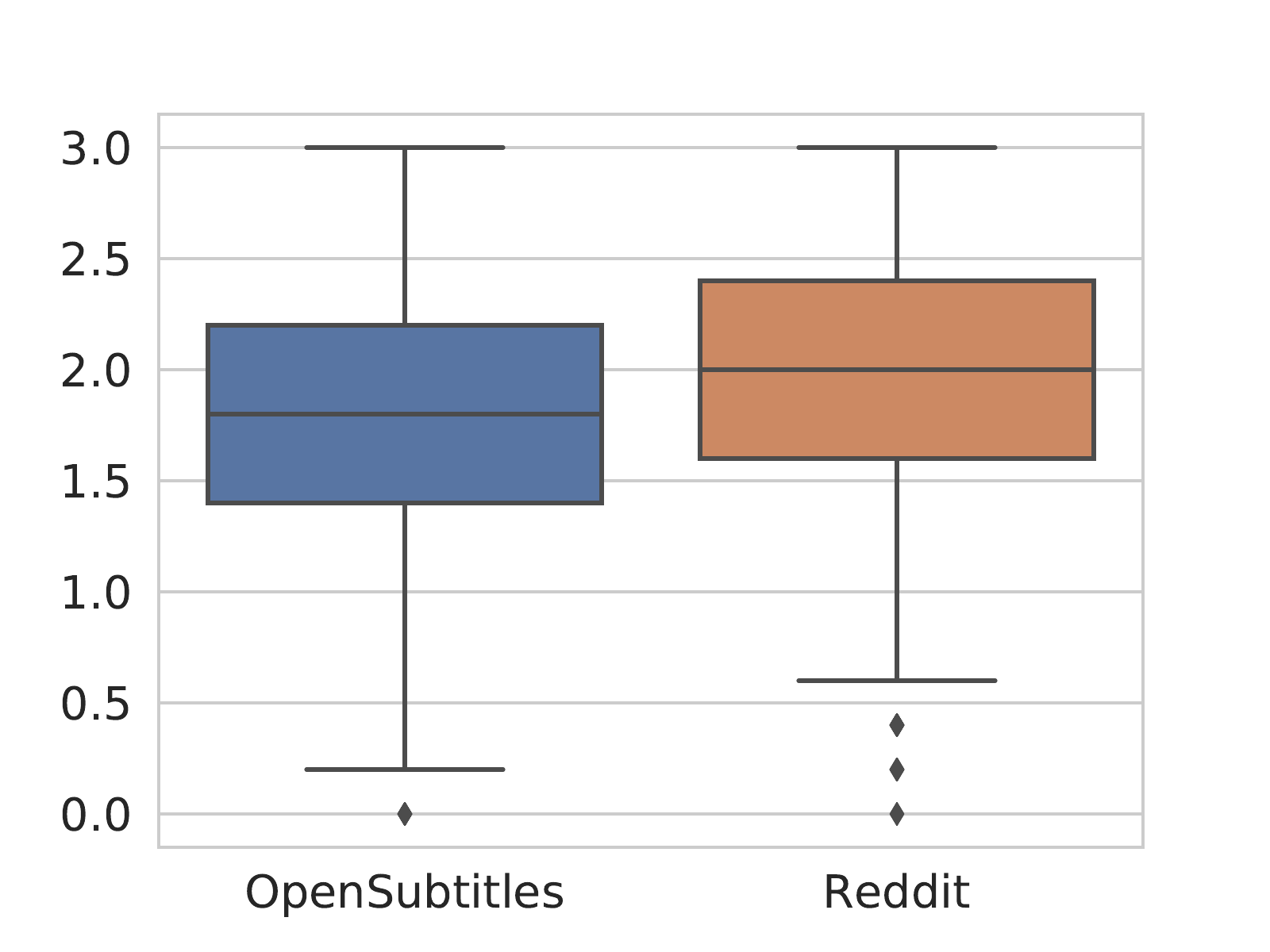}
\subcaption{Human MER}
\label{fig:dataset_comparison_human}
\endminipage
\minipage{.45\textwidth}
\includegraphics[width=\linewidth]{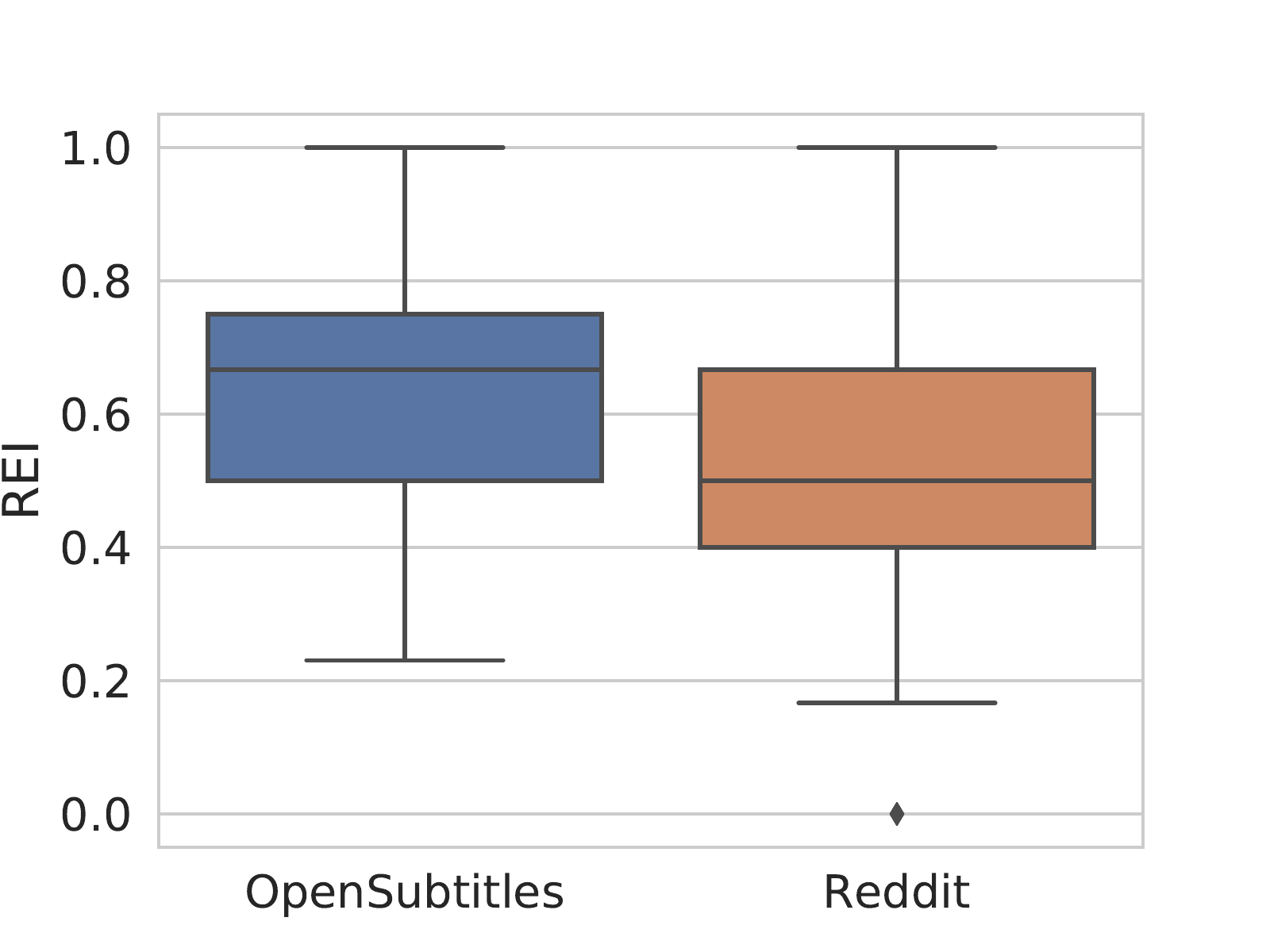}
\subcaption{REI}
\label{fig:dataset_comparison_sc_utt1}
\endminipage \hfill
\minipage{.45\textwidth}
\includegraphics[width=\linewidth]{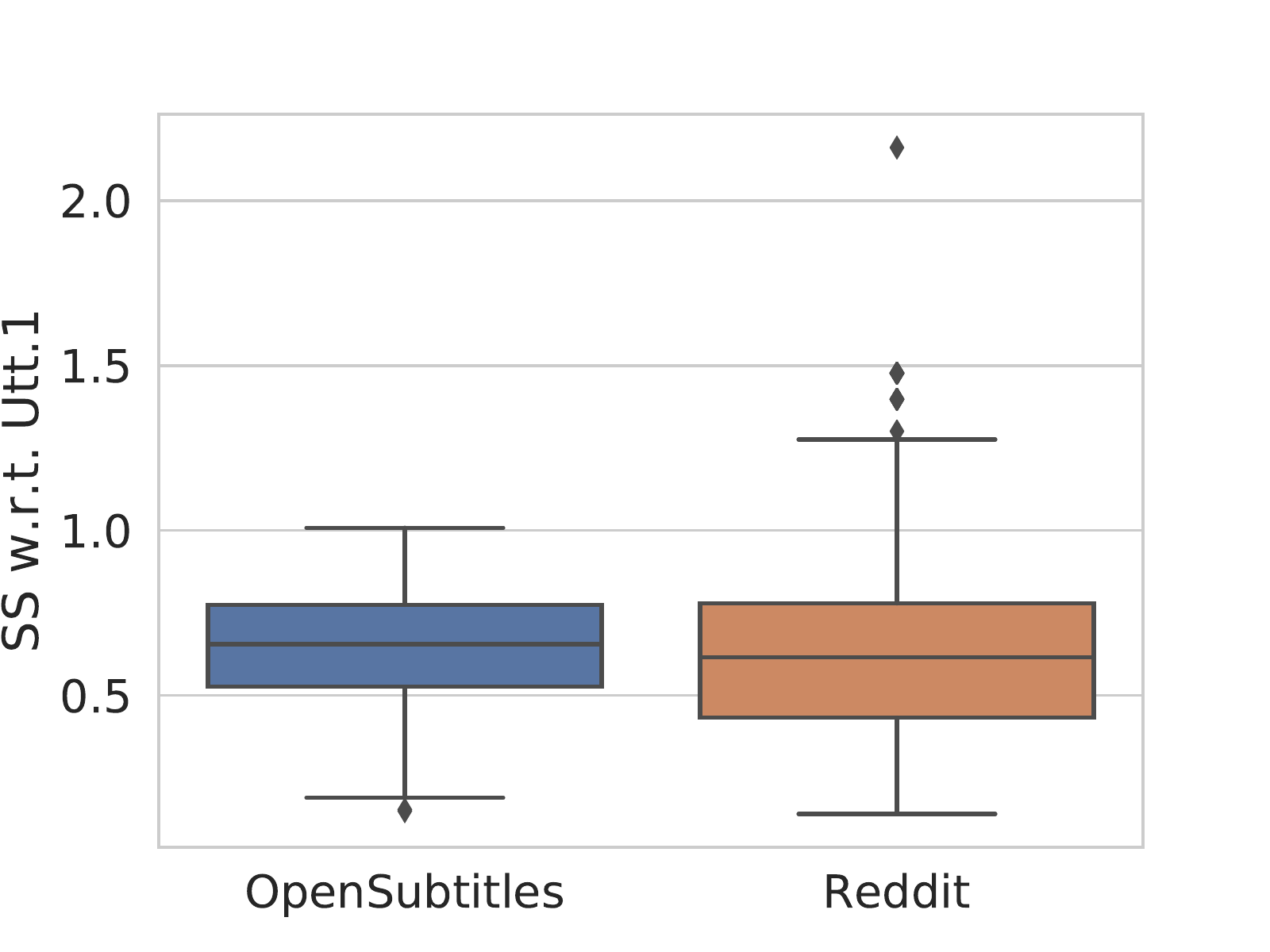}
\subcaption{Semantic Similarity w.r.t. Utt.1}
\label{fig:dataset_comparison_sc_utt2}
\endminipage
\minipage{.45\textwidth}
\includegraphics[width=\linewidth]{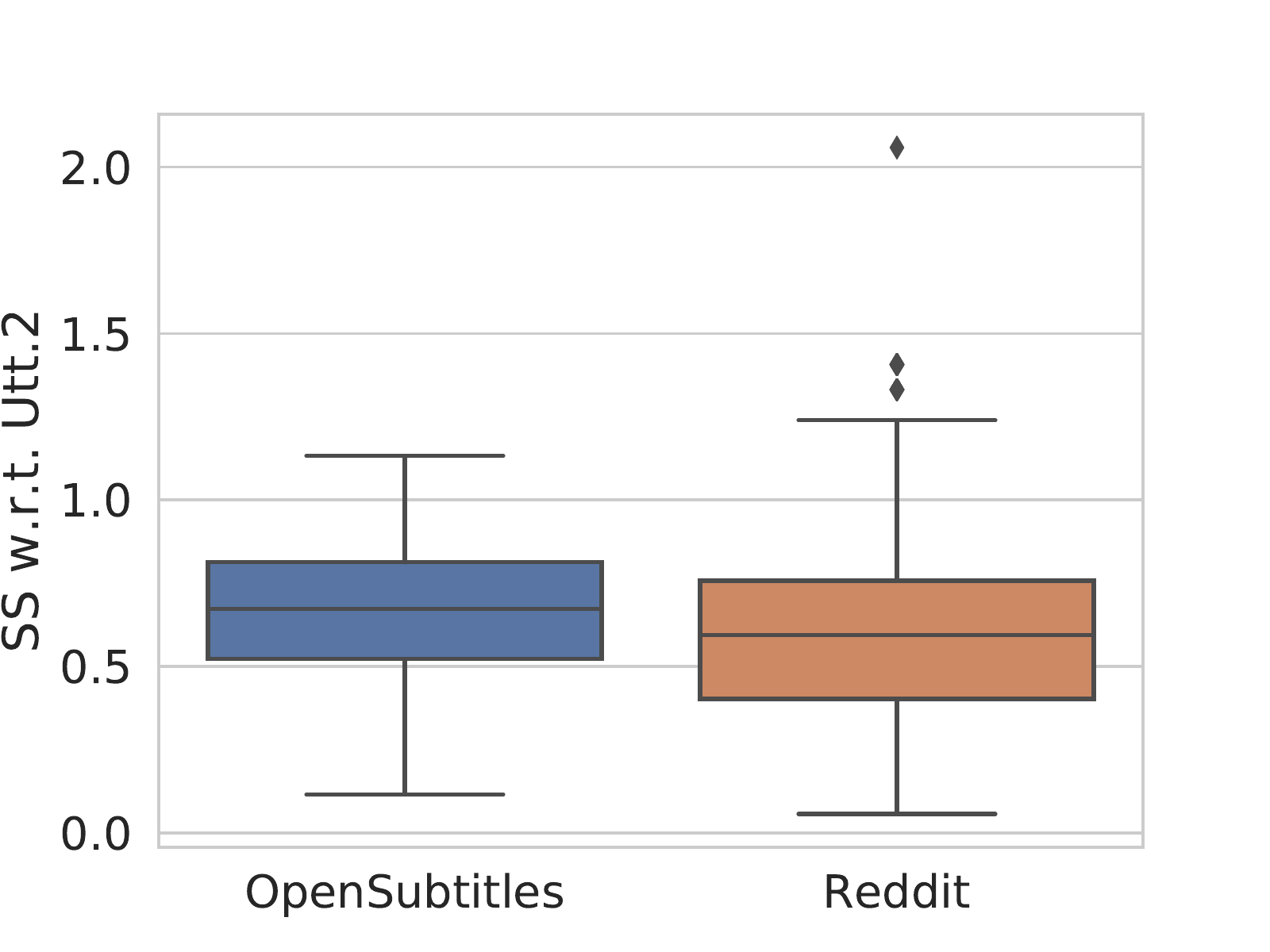}
\subcaption{Semantic Similarity w.r.t. Utt.2}
\label{fig:dataset_comparison_rei}
\endminipage
\caption{Box plots demonstrating the detailed comparison between OpenSubtitles and Reddit datasets. The metrics are calculated for all models in the cherry-picked data (150 samples for OpenSubtitles and 150 samples for Reddit). The results here complement what we found in Table~\ref{tab:dataset_comparison} in which only mean and standard deviation are reported per metric. 
}
\label{fig:dataset_comparison}
\end{figure*}

\end{document}